\documentclass{article}

\PassOptionsToPackage{numbers,square,comma,sort&compress}{natbib}
\usepackage[preprint]{neurips_2024}

\usepackage[utf8]{inputenc}
\usepackage[T1]{fontenc}
\usepackage{microtype}
\usepackage{xcolor}
\usepackage{graphicx}
\usepackage{amsmath,amssymb}
\usepackage{float}
\usepackage{booktabs}
\usepackage{threeparttable}
\usepackage{array,longtable}
\newlength{\suppwidth}
\newcommand{\ReserveSpace}[1]{\par\begingroup
  \dimen0=\pagegoal\advance\dimen0 by -\pagetotal
  \ifdim\dimen0<#1\relax\newpage\fi\endgroup}
\newcolumntype{L}[1]{>{\raggedright\arraybackslash}p{#1}}
\newcolumntype{C}[1]{>{\centering\arraybackslash}p{#1}}
\newcolumntype{R}[1]{>{\raggedleft\arraybackslash}p{#1}}
\providecommand{\NewStructureName}[1]{}
\providecommand{\AssignStructureRole}[2]{}
\usepackage[most]{tcolorbox}
\usepackage[hidelinks]{hyperref}

\title{A General Multimodal Probing Framework for Data Fusion and Prediction with Frozen Large Language Models: Evidence from Multimodal EHR Data}

\author{%
  Chengyuan Liu\thanks{Corresponding authors: Chengyuan Liu (\texttt{cjl6934@psu.edu}) and Guanting Chen (\texttt{guanting@unc.edu}).}\\
  Department of Statistics\\
  Pennsylvania State University\\
  University Park, PA, USA\\
  \texttt{cjl6934@psu.edu}
  \And
  Xinyue Zhang\\
  Department of Biostatistics\\
  University of North Carolina at Chapel Hill\\
  Chapel Hill, NC, USA\\
  \texttt{xinyuez@unc.edu}
  \AND
  Yao Li\\
  Department of Statistics and Operations Research\\
  University of North Carolina at Chapel Hill\\
  Chapel Hill, NC, USA\\
  \texttt{yaoli@unc.edu}
  \And
  Guanting Chen\footnotemark[1]\\
  Department of Statistics and Operations Research\\
  University of North Carolina at Chapel Hill\\
  Chapel Hill, NC, USA\\
  \texttt{guanting@unc.edu}
}

\begin{document}

\maketitle

\begin{abstract}
\normalfont\textbf{Objective:} The main objective of this paper is to propose a general framework for prediction based on different sources of multimodal data in the healthcare domain. We evaluated whether frozen medical large language model (LLM) representations can serve as a shared embedding space for multimodal primary diagnosis category prediction.\\[0.5em]
\normalfont\textbf{Materials and Methods:} We propose a backbone-flexible probing framework that serializes structured electronic health record (EHR) variables and leakage-pruned discharge-note sections into a shared frozen-LLM representation space and recovers primary diagnosis categories with linear probes. On a MIMIC-IV cohort of 13,645 admissions (seven diagnosis categories), we probed inputs of different modalities (Structured-only, Unstructured-only, and Combined) across transformer layers and six frozen backbones, compared against XGBoost and three text baselines (CAML, LAAT, PLM-ICD) under a matched split and context budget, and tested robustness across datasets on MIMIC-III.\\[0.5em]
\normalfont\textbf{Results:}
The Combined terminal-layer probe reached 88.86\% top-1 accuracy on MIMIC-IV, significantly outperforming CAML and PLM-ICD and statistically comparable to LAAT; Combined accuracy ranged from 87.69\% to 89.25\% across six backbones. On MIMIC-III, in-domain Combined probing reached 74.18\% accuracy and significantly exceeded all three baselines, and in a transfer-learning test a 2.8M-parameter adapter on frozen MIMIC-IV representations reached 92.51\% accuracy, exceeding adapted LAAT (85.54\%).\\[0.5em]
\normalfont\textbf{Discussion:} Frozen LLM embeddings unify structured and narrative EHR information: the Combined configuration was strongest in every setting, and its representations transferred across datasets through a compact adapter.\\[0.5em]
\normalfont\textbf{Conclusion:} Multimodal probing of frozen LLM representations provides a practical, backbone-flexible approach for studying EHR modalities and adapting clinical representations across datasets.
\end{abstract}



\section{Introduction}

\subsection{Background and Significance}

Electronic health records (EHRs) combine complementary clinical signals from structured measurements, medications, demographics, and narrative documentation. Although multimodal EHR data are increasingly available, and large language models (LLMs) are increasingly applied to these data, it remains unclear how these heterogeneous sources should be organized and analyzed so that additional modalities translate into reliable predictive gains rather than added noise, redundancy, leakage, or task-specific complexity. In particular, we ask whether structured variables and clinical narratives can be represented within a common framework that consistently reveals their individual and joint diagnostic value across models and datasets. We study this question using primary ICD-derived diagnosis categories as a controlled label space~\cite{who1992international}.

Prior work has established that heterogeneous EHR sources can be extracted, aligned, and combined for clinical prediction: through reproducible data pipelines~\cite{adiba2026multimodal,dou2024quick}, through task-specific fusion architectures~\cite{xu2019multimodal,liu2022treeman,liu2022multimodal,merchant2024ensemble,thao2024medfuse}, and, more recently, through language-model approaches that serialize structured records or place them alongside clinical notes in a unified LLM context~\cite{ruan2025ptransformer,lee2025clinical,zhu2026clinicrealm,hegselmann2025large}.

However, an important methodological gap remains. General-purpose data pipelines typically stop at producing analysis-ready datasets, whereas predictive systems often remain tied to task-specific encoders, learned fusion modules, backbone fine-tuning, or direct generative prompting. Existing language-model-based approaches have largely occupied one of two settings: they either apply frozen encoders to serialized structured EHR records without incorporating genuine clinician-authored narratives~\cite{hegselmann2025large,lee2025clinical}, or they combine structured and narrative data through specialized multimodal architectures with substantial task-specific training~\cite{xu2019multimodal, liu2022treeman, thao2024medfuse, engelke2025fhir}. As a result, observed predictive gains do not directly reveal whether the additional modality contributes useful information within a common representation, independently of \textbf{increased trainable capacity} or \textbf{architectural complexity}. A lightweight, controlled, and reusable framework for analyzing the individual and joint predictive value of structured and narrative EHR data therefore remains underexplored.

Constructing such a framework is nontrivial. Structured EHR variables are sparse, irregularly sampled, duplicated, unit-dependent, and affected by informative missingness, whereas clinical narratives contain copied-forward text, heterogeneous section structures, redundant structured information, and potentially explicit statements of the prediction target. Joint modeling introduces additional challenges, including temporal misalignment, unequal allocation of the context window, duplicated evidence across modalities, and dominance of long narrative inputs over shorter structured summaries. Consequently, naïvely concatenating the two sources may fail to exploit their complementary information, obscure the contribution of each modality, or produce misleading gains through label leakage.

We address these challenges with a leakage-mitigated, backbone-flexible probing framework that places structured EHR variables and genuine clinical narratives in a shared frozen-LLM representation space. Structured variables are summarized and serialized into a standardized clinical narrative; discharge-note sections are extracted and diagnosis-bearing tail sections removed. Structured-only, Unstructured-only, and Combined inputs are then evaluated with the same cohort, prompt constraints, frozen LLM, and linear-probe specification, so that only the input modality changes. Unlike prior multimodal systems that rely on modality-specific encoders or learned fusion modules, this matched design isolates the incremental diagnostic information contributed by each modality, and the same frozen representations support layer-wise analysis, in-domain replication on MIMIC-III, and compact adaptation across datasets. Our experiments show that the Combined configuration consistently produces the strongest terminal-layer diagnostic representations across all six frozen backbones. Our aim is therefore not to propose a new task-specific multimodal architecture, but to provide a controlled and reusable methodology for constructing, analyzing, and realigning multimodal EHR representations through a shared frozen LLM.

\section{Materials and Methods}
\begin{figure*}[t]
\centering
\includegraphics[width=0.98\textwidth]{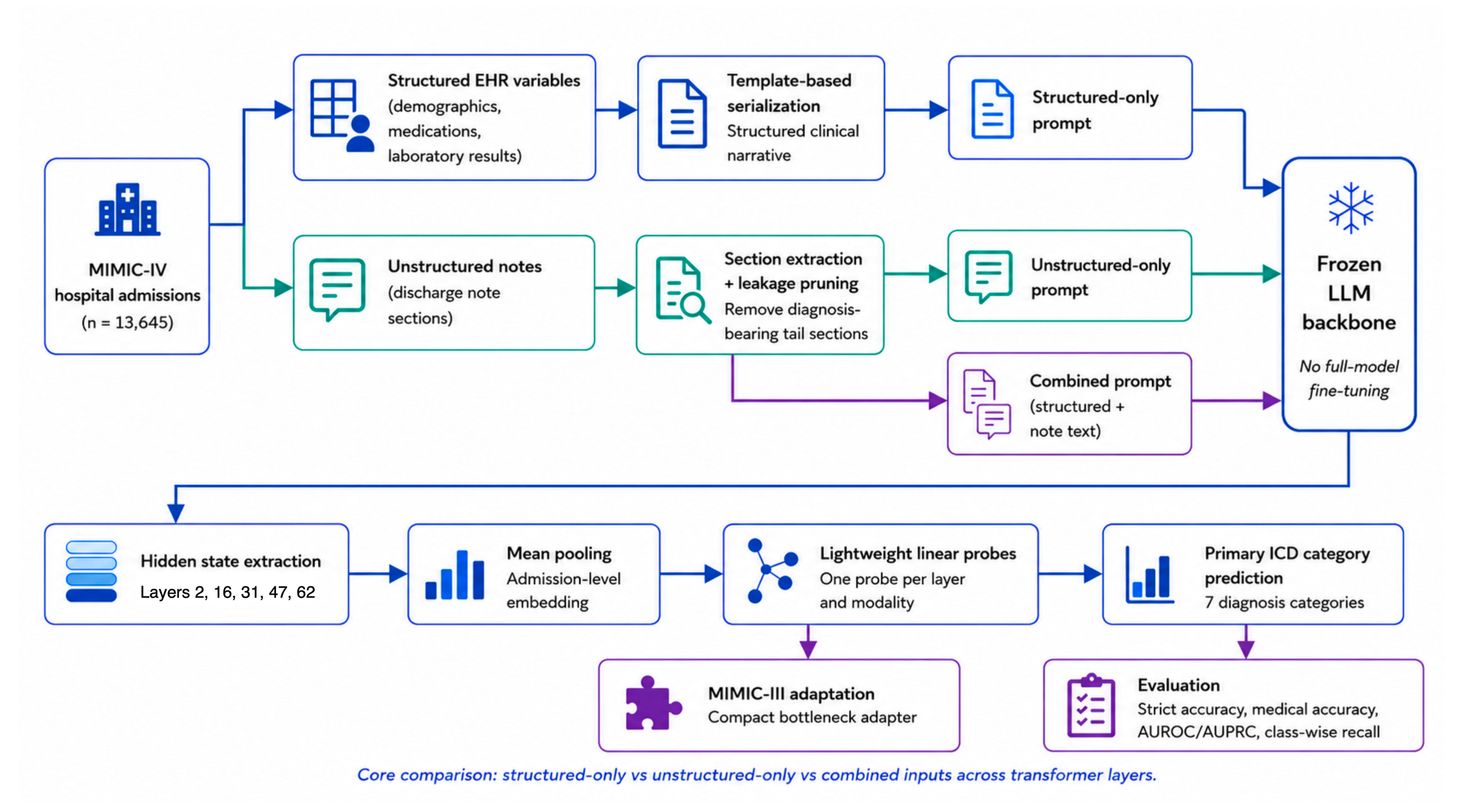}
\caption{Overview of the multimodal probing framework. Structured and unstructured EHR inputs are converted to text prompts, encoded by a frozen LLM backbone, and evaluated with layer-wise linear probes under Structured-only, Unstructured-only, and Combined configurations. The conceptual pipeline is independent of any specific LLM checkpoint.}
\label{fig:method_framework}
\end{figure*}

As illustrated in Figure~\ref{fig:method_framework}, each hospitalization is one sample with two types of data, \textbf{structured} variables and \textbf{unstructured} discharge-note text, both serialized into text so that the same frozen backbone encodes them without a task-specific fusion model. We evaluate three input configurations, each corresponding to an input path in Figure~\ref{fig:method_framework}: \textbf{Structured-only}, the structured variables (demographics, medications, and laboratory results) serialized into a clinical narrative; \textbf{Unstructured-only}, the leakage-pruned discharge-note sections (chief complaint, history of present illness, physical exam, brief hospital course, and pertinent results); and \textbf{Combined}, the structured narrative and the discharge-note sections concatenated into a single prompt.

\subsection{Data Source and Task Definition}

We used MIMIC-IV as the primary dataset and MIMIC-III for in-domain replication and cross-dataset transfer~\cite{mimiciv31,mimiciii14}. The MIMIC-IV cohort comprised 13,645 admissions from the 10 most frequent primary ICD-10 codes, consolidated into seven clinically meaningful diagnosis categories~\cite{rao20252025,rangaswami2019cardiorenal,zhang2021construction}, and was split 70\%/15\%/15\% into 9,556 training, 2,042 validation, and 2,047 test admissions (Supplementary Table S1; code-level grouping in Supplementary Table S3). This label space and split were used across all probing and baseline analyses.

We defined three input configurations $c \in \mathcal{C} = \{\text{str}, \text{txt}, \text{combined}\}$. Structured variables were mapped to a clinical narrative through a template $T_{\text{str}}$, and discharge notes were reduced to clinically relevant content through a section-extraction function $T_{\text{txt}}$:
\begin{tcolorbox}[breakable, colback=gray!10, colframe=gray!70, title=Example input, fonttitle=\bfseries\small,
  width=\columnwidth, boxsep=2pt, left=3pt, right=3pt, top=2pt, bottom=2pt]
\footnotesize\sloppy
\textbf{Structured ($T_{\text{str}}$):} Patient: [age]-year-old [gender] $\mid$ Medications: \ldots $\mid$ Laboratory results: \ldots

\medskip
\textbf{Unstructured ($T_{\text{txt}}$):} Chief Complaint: \ldots $\mid$ History of Present Illness: \ldots $\mid$ Pertinent Results: \ldots
\end{tcolorbox}
The Structured-only and Unstructured-only configurations used $T_{\text{str}}$ and $T_{\text{txt}}$ respectively, while the Combined configuration concatenated both sources under a shared template, $P_{\text{combined}} = T_{\text{str}} \oplus T_{\text{txt}}$, where $\oplus$ denotes concatenation with a fixed delimiter. Every configuration ended with the same diagnostic query to standardize the prediction context across modalities.

To mitigate label leakage, we removed discharge-tail sections that state the diagnosis directly (e.g., discharge diagnosis and final impression), retaining only sections that describe the clinical course and findings. Residual diagnosis-bearing text would otherwise inflate apparent separability. The same pruning was applied identically across configurations and to all baseline comparisons. Because the retained sections can still mention diagnoses and the structured variables span the full stay, this construction mitigates rather than eliminates leakage, and the task is best read as retrospective coding assistance at discharge. To test how much the probe relies on explicit diagnosis statements, we additionally masked every diagnosis-revealing keyword for the seven categories (e.g., ``sepsis'', ``NSTEMI'', ``heart failure''; 96.5\% of admissions, $\sim$93{,}000 occurrences), re-extracted the embeddings, and retrained the probes.

Full serialization templates, variable-selection rules (including medication and laboratory-test limits and abnormal-value prioritization), section-boundary detection, and resulting input-length statistics are provided in the Supplementary Methods.

\subsection{Frozen-LLM Probing Framework}
The framework treats the LLM as a replaceable frozen representation function. Because all backbone parameters stay fixed, supervision enters only through the probe. We instantiated six frozen decoder backbones spanning 8B--70B parameters, two architecture families, and general-domain or medical training. The primary backbone for all confirmatory analyses (MedGemma-27B) was selected as the backbone with the highest validation-set Combined accuracy, without reference to test performance; MedFound-Llama3-8B serves as the sensitivity backbone. The primary depth analysis extracted five approximately evenly spaced layers from the 62-layer primary backbone (layers 2, 16, 31, 47, and 62); cross-backbone robustness experiments used each model's terminal representation while holding the cohort, split, prompts, pooling strategy, and probe specification fixed. Checkpoint identities and representation dimensions are given in the Supplementary Methods.

For backbone $b$, admission $i$, input configuration $c \in \mathcal{C}$, and selected layer $l \in \mathcal{L}_b$, we mean-pooled token-level hidden states over non-padding tokens to obtain an admission-level representation $h_{i,b}^{(c,l)} \in \mathbb{R}^{d_b}$, where representation width $d_b$ and terminal depth depend on the backbone. We then trained a linear probe to map each representation to $M=7$ diagnosis-category scores:
{
\[
z_{i,b}^{(c,l)} = W_b^{(c,l)} h_{i,b}^{(c,l)} + b_b^{(c,l)},
\]
}
where $z_{i,b}^{(c,l)} \in \mathbb{R}^{M}$ are the class logits and $W_b^{(c,l)} \in \mathbb{R}^{M \times d_b}$, $b_b^{(c,l)} \in \mathbb{R}^{M}$ are the probe parameters, the only parameters optimized (cross-entropy loss); the backbone remained frozen. We restricted the probe to a linear map so that recoverable diagnostic signal is attributable to the frozen representations rather than to probe capacity~\cite{alain2016understanding,hewitt2019designing}.

Probe training settings and optimization details are provided in the Supplementary Methods, and more details in Supplementary Tables S13--S14. 

\subsection{Baseline Methods}
An effective and robust pipeline should satisfy two criteria: first, it should outperform baseline methods across different data modalities; second, its performance should improve with the inclusion of additional modalities. To assess the first criterion, we introduce the following baseline methods for comparison: a conventional non-LLM model for \textbf{structured EHR data} and three task-specific neural architectures for clinical text (\textbf{unstructured EHR data}).


\paragraph{Structured Baseline (XGBoost):}
We trained an XGBoost classifier on laboratory, vital-sign, and medication-derived features~\cite{chen2016xgboost}, summarizing longitudinal variables into fixed-length features while retaining missingness for sparsity-aware splitting.

\paragraph{Unstructured Baselines:}
We evaluated three established ICD-coding architectures using the same admission-level split and a text budget matched to PLM-ICD: CAML, a convolutional encoder with label-specific attention~\cite{mullenbach2018explainable}; LAAT, a bidirectional LSTM with label-specific attention~\cite{vu2020label}; and PLM-ICD, a pretrained BERT encoder with label attention over long-note segments~\cite{huang2022plm}.

We found no directly comparable external baseline for the Combined configuration. We therefore evaluated the text baselines on their native note-only inputs and, as a same-input ablation, retrained them on the serialized Combined text used by the probe.


We did not identify a reproducible end-to-end multimodal comparator under the same split, inputs, and prediction target. The closest prior ICD-prediction system~\cite{xu2019multimodal} has no public implementation and does not fully specify its structured-variable selection. Widely used structured-EHR models instead target outcomes such as mortality, readmission, or length of stay under a patient-level framing~\cite{rajkomar2018scalable,harutyunyan2019multitask}, while recent fusion architectures~\cite{thao2024medfuse,lyu2023multimodal,merchant2024ensemble} use prediction targets and training regimes that differ from ours. Full preprocessing and training details, baseline adaptations, and the rationale for comparator selection are provided in the Supplementary Methods.

\subsection{Evaluation and Statistical Analysis}

We report two accuracy metrics over the seven categories. \emph{Strict accuracy} is exact-match top-1 accuracy and is the primary metric. \emph{Medical accuracy} is a clinically relaxed secondary metric that additionally credits exactly two clinically related cross-category pairs: Cardiovascular Event with Hypertensive Heart Failure, and Acute Kidney Failure with Hypertensive Heart Failure (cardiorenal syndrome)~\cite{rangaswami2019cardiorenal}; for example, predicting Hypertensive Heart Failure for an acute myocardial infarction is counted as medically related but strictly incorrect, and every other confusion is an error under both metrics. All models, including the text baselines trained on the 10 raw ICD-10 codes, are scored in this same seven-category space (Supplementary Table~S2). We additionally report per-class recall, macro accuracy, and, when class scores are available, macro AUROC and macro AUPRC.

For paired comparisons on the same admissions, we used exact McNemar tests for strict accuracy and paired bootstrap resampling for differences in continuous metrics, with Benjamini-Hochberg false discovery rate control across multiple planned tests. We interpret layer-wise trends rather than small differences between individual layers. Detailed implementation settings, the full soft-matching specification, and complete statistical procedures are provided in the Supplementary Methods.


\section{Results}

We conduct experiments that highlight the advantages of the proposed pipeline. Overall, our pipeline matched or outperformed all competing methods across datasets and experimental settings, effectively leveraged additional data modalities, and demonstrated robust and transferable learning. More specifically, we conduct experiments on its robustness across transformer layers, input modalities, and different frozen LLM backbones (Table~\ref{tab:probing}; Supplementary Table S12), its advantage over task-specific baselines on MIMIC-IV (Table~\ref{tab:plmicd_matched}), its in-domain replication on a second dataset (MIMIC-III; Table~\ref{tab:mimic3_baselines}), and its cross-dataset transferability (Table~\ref{tab:mimic3_transfer_baselines}). More detailed experimental results can be found in the Supplementary Materials.

\subsection{Probing Pipeline's Performance}

\begin{figure}[htbp]
\centering
\includegraphics[width=\columnwidth]{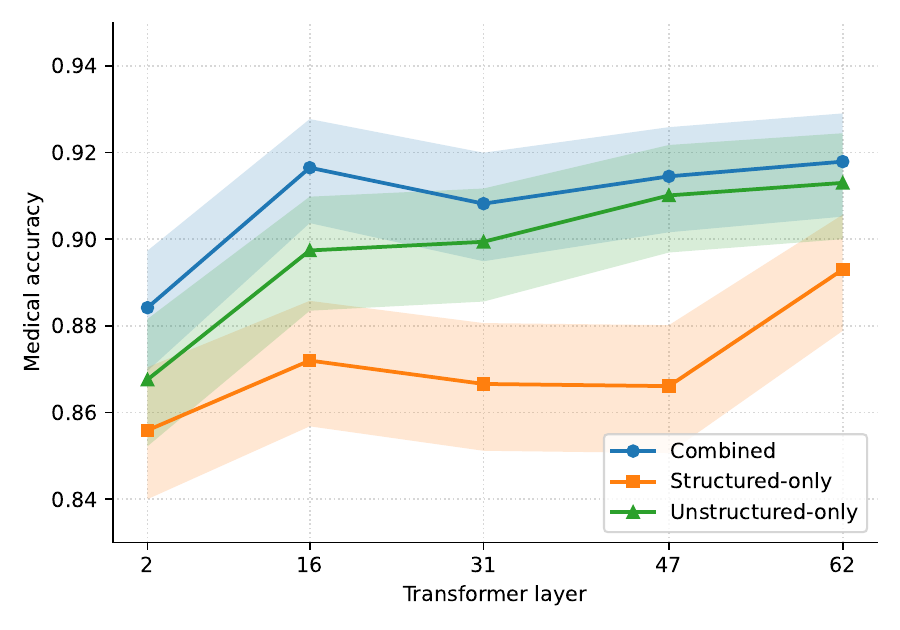}
\caption{Layer-wise medical accuracy for linear probes trained on frozen MedGemma-27B embeddings across input configurations and data modality. Shaded bands denote 95\% Wilson score intervals over the $n=2{,}047$ canonical test admissions.}
\label{fig:layerwise_accuracy}
\end{figure}

Figure~\ref{fig:layerwise_accuracy} and Table~\ref{tab:probing} present linear probing medical accuracy across transformer layers and data configurations. At Layer 62, the Combined configuration reached 91.79\% medical accuracy, compared with 89.30\% for Structured-only and 91.30\% for Unstructured-only on the same canonical test admissions. Paired analyses showed the clearest and most consistent gains over Structured-only, while the margin over Unstructured-only was smaller; full multi-metric and pairwise probing results are reported in Supplementary Tables S4--S5. Medical accuracy was already high at Layer 2 (88.42\% for Combined) and rose modestly and non-monotonically with depth, with every configuration peaking at the terminal layer. This profile is flatter than the steep shallow-to-intermediate rise of the MedFound sensitivity backbone: where diagnostic information consolidates along depth is backbone-dependent, whereas the terminal-layer advantage is not.

The probe does not depend on explicit diagnosis mentions. Masking every diagnosis keyword left Combined accuracy unchanged (92.18\% vs.\ 91.79\% medical accuracy, McNemar $p=0.37$) and Unstructured-only medical accuracy unchanged (90.96\% vs.\ 91.30\%, $p=0.47$; strict $-1.03$ points, not significant after correction), indicating that the recoverable signal lies in the surrounding clinical context (Supplementary Table S15).

\begin{table}[htbp]
\centering
\caption{Linear probing medical accuracy across transformer layers and data modality on the MIMIC-IV canonical split (MedGemma-27B primary backbone).}
\label{tab:probing}
\begin{threeparttable}
\small
\setlength{\tabcolsep}{4pt}
\begin{tabular}{lccc}
\hline
\textbf{Layer} & \textbf{Combined} & \textbf{Structured} & \textbf{Unstructured} \\
\hline
2 & 0.8842 & 0.8559 & 0.8676 \\
16 & 0.9165 & 0.8720 & 0.8974 \\
31 & 0.9082 & 0.8666 & 0.8994 \\
47 & 0.9145 & 0.8661 & 0.9101 \\
62 & \textbf{0.9179} & \textbf{0.8930} & \textbf{0.9130} \\
\hline
\end{tabular}
\begin{tablenotes}
\small
\item Models were trained and evaluated on seven merged diagnosis categories. Best performance per configuration is bold. Medical accuracy credits exact matches and clinically related predictions under the fixed soft-matching rule described in the Evaluation metrics section.
\end{tablenotes}
\end{threeparttable}
\end{table}

\subsection{Consistency Across LLM Backbones}

We applied the same terminal-layer probing protocol to six frozen LLM backbones spanning 8B to 70B parameters, general-domain and medical training, and the Llama and Gemma model families. Despite these differences, Combined performance remained within a narrow range, with strict accuracy of 87.69--89.25\% and medical accuracy of 91.45--92.82\% (Supplementary Table S12), and for every backbone the Combined configuration achieved the highest point estimate among the three input configurations. Gains over Structured-only were consistently larger than gains over Unstructured-only. The cross-backbone values are descriptive, and we do not test between backbones.

\subsection{Per-Diagnosis Performance Comparison}

\begin{figure}[htbp]
\centering
\includegraphics[width=\columnwidth]{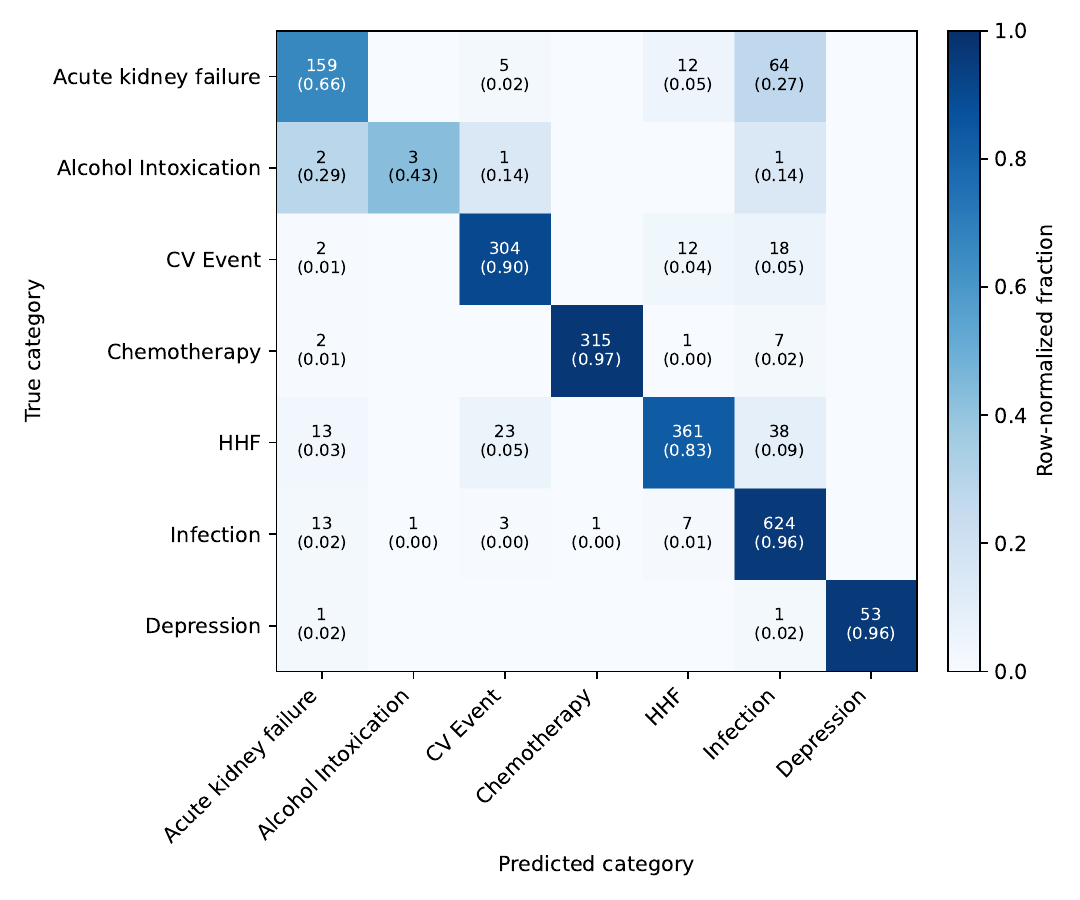}
\caption{Confusion matrix of the primary Combined Layer-62 probe (MedGemma-27B) on the canonical MIMIC-IV test set ($n=2{,}047$), with row-normalized fractions in parentheses. Cardiovascular Event, Hypertensive Heart Failure (HHF), and Infection-Related are merged categories; the remaining rows are single ICD-10 codes.}
\label{fig:confusion_matrix}
\end{figure}

Under exact matching, strict recall for the Combined Layer-62 probe was highest for chemotherapy encounters, major depression, infection-related diagnoses, and cardiovascular events, and lowest for acute kidney failure (Supplementary Table S6; the alcohol-intoxication category had only seven test cases, so its estimate is unstable). Errors concentrated in two clinically overlapping categories (Figure~\ref{fig:confusion_matrix}): acute kidney failure was most often assigned to Infection-Related (64/240) or Hypertensive Heart Failure (12/240), and Hypertensive Heart Failure to Infection-Related (38/435) or Cardiovascular Event (23/435). Macro precision was 0.890, macro recall 0.817, and macro-F1 0.845 against 88.86\% overall strict accuracy, reflecting class imbalance.

\subsection{Comparison with Baselines}

\paragraph{Structured Baseline:}
The XGBoost structured baseline reached 81.60\% medical accuracy on structured EHR features, whereas the Structured-only MedGemma probe at Layer 62 reached 89.30\%, a 7.70-point improvement. Serializing tabular variables into natural language therefore recovers structured diagnostic signal beyond a conventional gradient-boosted baseline, without updating the backbone.

\paragraph{Unstructured Baselines:}
Table~\ref{tab:plmicd_matched} compares the Combined probe with CAML, LAAT, and PLM-ICD under the same canonical split, leakage pruning, and $\sim$3,000-character budget.

\begin{table}[htbp]
\centering
\caption{Comparison with task-specific text baselines on MIMIC-IV under a matched split and context budget.}
\label{tab:plmicd_matched}
\begin{minipage}{\columnwidth}
\centering
\begin{threeparttable}
\footnotesize
\setlength{\tabcolsep}{3pt}
\begin{tabular}{@{}lcccc@{}}
\hline
\textbf{Model} & \textbf{Strict} & \textbf{Medical} & \textbf{AUROC} & \textbf{AUPRC} \\
\hline
Combined probe & \textbf{0.8886} & \textbf{0.9179} & \textbf{0.9880} & \textbf{0.9154} \\
PLM-ICD & 0.8647 & 0.9062 & 0.9432 & 0.7919 \\
CAML & 0.8647 & 0.9062 & 0.9368 & 0.7684 \\
LAAT & 0.8837 & 0.9174 & 0.9865 & 0.8681 \\
\hline
\end{tabular}
\begin{tablenotes}[flushleft]
\footnotesize
\item Combined denotes the primary backbone (MedGemma-27B) Layer-62 probe; CAML and LAAT use the pre-specified seed 42. Across three seeds, CAML reached $85.98\%\pm0.49\%$ strict / $90.39\%\pm0.22\%$ medical and LAAT $88.73\%\pm0.31\%$ / $92.25\%\pm0.45\%$. All models share the canonical split, leakage pruning, $\sim$3,000-character budget, and seven-category evaluation space.
\item The Combined probe receives serialized structured variables in addition to note text within the same character budget, whereas the text baselines receive note text only; this comparison therefore reflects both representation and input differences.
\item Macro AUROC/AUPRC are one-vs-rest averages over seven categories.
\end{tablenotes}
\end{threeparttable}
\end{minipage}
\end{table}


The Combined probe achieved the highest strict (88.86\%) and medical accuracy (91.79\%) among all text models, together with the highest macro AUROC and AUPRC. It significantly exceeded PLM-ICD and CAML in strict accuracy ($+2.39$ points for both, BH-adjusted $p=0.0063$) and on every class-balanced metric (macro-F1, macro AUROC, macro AUPRC; all $p\leq0.0028$), while the medical accuracy margin ($+1.17$ points) did not reach significance ($p=0.12$). Against LAAT, the strongest baseline, the probe was statistically comparable in strict and medical accuracy ($\Delta=+0.49$, $p=0.64$; $\Delta=+0.05$, $p=0.97$) and in macro AUROC, with higher macro AUPRC (0.9154 vs 0.8681, $p=0.036$). The MedFound-Llama3-8B sensitivity analysis (87.69\%/91.45\%) remained competitive in strict and medical accuracy and significantly exceeded CAML and PLM-ICD on class-balanced metrics (Supplementary Table S17).

The Combined input is itself a product of the framework: the serialization step converts tabular variables into text that the baselines cannot consume in their native form. Giving the baselines this same serialized text therefore turns the comparison into an ablation of two framework components, the serialized input and the frozen-LLM representation. We retrained all three baselines on exactly the Combined text consumed by the probe. On MIMIC-IV, the added structured text changed the baselines by less than half a point (CAML 86.47$\rightarrow$85.93\%, PLM-ICD 86.47$\rightarrow$86.91\%, LAAT 88.37$\rightarrow$88.72\% strict), and the Combined probe still significantly exceeded same-input CAML and PLM-ICD (strict BH-adjusted $p\leq0.030$), matched same-input LAAT in accuracy, and retained higher macro AUROC and AUPRC than all three. On MIMIC-III, the serialized text raised every baseline substantially (CAML 57.03$\rightarrow$66.82\%, LAAT 65.25$\rightarrow$70.57\%, PLM-ICD 63.88$\rightarrow$71.84\%), yet the same-input probe (74.18\%) still significantly exceeded all three (BH-adjusted $p\leq0.023$). Both components therefore contribute: the serialized input adds information that conventional encoders can use, and the frozen-LLM representation extracts more from that input than the trained encoders do (Supplementary Table S18).

Finally, a TF-IDF + logistic-regression model on the same Combined text reached 88.72\% strict accuracy, indistinguishable from the probe, but trailed it by 8.3 points in macro-F1 ($p=0.024$). Lexical features thus set the floor of overall accuracy on this task, while representation quality sets the ceiling on class-balanced performance, consistent with the masking result above.

\subsection{Replication and Cross-Dataset Adaptation on MIMIC-III}

We then trained the probe in our pipeline on the MIMIC-III cohort to evaluate reproducibility and transferability. In a separate transfer experiment, a compact adapter allowed the probe trained on MIMIC-IV to predict MIMIC-III diagnoses using limited labeled MIMIC-III training data. The first experiment included 13,137 admissions covering the 10 most frequent ICD-9 codes, consolidated into seven diagnosis categories using the same input preprocessing as for MIMIC-IV (Supplementary Table S7). Both datasets originate from the same institution but differ in documentation style and coding system, so these results support transfer across datasets within one institution.

\subsubsection{In-Domain Replication on MIMIC-III}

Retrained in-domain, the layer-wise pattern recapitulated the MIMIC-IV results, with the Combined configuration again best at the terminal layer (Supplementary Table S9--S11). Table~\ref{tab:mimic3_baselines} mirrors the MIMIC-IV baseline comparison on the MIMIC-III test set ($n=1{,}971$).

\begin{table}[htbp]
\centering
\caption{Comparison with task-specific text baselines on MIMIC-III (in-domain, ICD-9) under a matched split and context budget.}
\label{tab:mimic3_baselines}
\begin{minipage}{\columnwidth}
\centering
\begin{threeparttable}
\footnotesize
\setlength{\tabcolsep}{3pt}
\begin{tabular}{@{}lcccc@{}}
\hline
\textbf{Model} & \textbf{Strict} & \textbf{Medical} & \textbf{AUROC} & \textbf{AUPRC} \\
\hline
Combined probe & \textbf{0.7418} & \textbf{0.8884} & \textbf{0.9467} & \textbf{0.7082} \\
PLM-ICD & 0.6388 & 0.7986 & 0.8592 & 0.5226 \\
CAML & 0.5703 & 0.7362 & 0.7824 & 0.3788 \\
LAAT & 0.6525 & 0.8148 & 0.8870 & 0.5582 \\
\hline
\end{tabular}
\begin{tablenotes}[flushleft]
\footnotesize
\item Combined denotes the MedGemma-27B Layer-62 probe; CAML and LAAT use seed 42, PLM-ICD a single seed-42 run. All models share the MIMIC-III split, leakage pruning, $\sim$3,000-character budget, and seven-category space.
\item Combined significantly exceeded all three baselines on strict and medical accuracy (all BH-adjusted $p<0.001$); full gains and CIs are in Supplementary Table S20.
\item Macro AUROC/AUPRC are one-vs-rest averages over seven categories.
\end{tablenotes}
\end{threeparttable}
\end{minipage}
\end{table}

Combined probing reached 74.18\% strict and 88.84\% medical accuracy and significantly exceeded LAAT (65.25\%; 81.48\% medical), PLM-ICD (63.88\%; 79.86\%), and CAML (57.03\%; 73.62\%), all BH-adjusted $p<0.001$ (Supplementary Table S19). LAAT's parity with the probe on MIMIC-IV did not persist here: the fully trained encoders degraded sharply on this cohort, whereas the probe on frozen representations retained most of its performance.

\subsubsection{Cross-Dataset Representation Transfer}

Frozen MIMIC-IV representations also transferred to MIMIC-III when target labels were limited. For cross-dataset transfer, five mappable MIMIC-III source diagnoses were mapped onto three MIMIC-IV target categories (Supplementary Table S8), giving 10,321 mapped admissions and a fixed three-category test set ($n=1{,}549$). Applying the frozen MIMIC-IV probe directly to MIMIC-III representations degraded accuracy to 50.1--88.7\% across modalities, a lower bound that quantifies documentation- and coding-induced shift (Supplementary Table S22).

This shift was corrected with a bottleneck adapter of approximately 2.8M parameters trained on target-domain embeddings while both the source backbone and the MIMIC-IV probe stayed frozen. For admission $i$ with Combined representation $h_i \in \mathbb{R}^{5376}$, the adapted representation is

\begin{align}
u_i &= \mathrm{ReLU}(W_{\mathrm{down}} h_i), \\
\hat{h}_i &= \mathrm{LayerNorm}(h_i + W_{\mathrm{up}}u_i).
\end{align}

where $W_{\text{down}} \in \mathbb{R}^{256 \times 5376}$ and $W_{\text{up}} \in \mathbb{R}^{5376 \times 256}$ are the only optimized parameters ($W_{\text{up}}$ zero-initialized; Supplementary Methods). With 100\% target supervision, Table~\ref{tab:mimic3_transfer_baselines} compares the adapted Combined probe against adapted LAAT, the strongest adaptable text baseline, on the same mapped test set; adapted PLM-ICD and the non-comparability of the in-domain rows are noted in the table.

\begin{table}[htbp]
\centering
\caption{Cross-dataset MIMIC-IV$\rightarrow$MIMIC-III transfer at full target supervision.}
\label{tab:mimic3_transfer_baselines}
\begin{minipage}{\columnwidth}
\centering
\begin{threeparttable}
\footnotesize
\setlength{\tabcolsep}{3pt}
\begin{tabular}{@{}lcccc@{}}
\hline
\textbf{Model} & \textbf{Strict} & \textbf{Medical} & \textbf{AUROC} & \textbf{AUPRC} \\
\hline
Combined probe + adapter & \textbf{0.9251} & \textbf{0.9251} & \textbf{0.9617} & \textbf{0.8106} \\
LAAT, adapted & 0.8554 & 0.8554 & 0.8819 & 0.6711 \\
\hline
\end{tabular}
\begin{tablenotes}[flushleft]
\footnotesize
\item Both rows are evaluated on the fixed three-category mapped transfer test set ($n=1{,}549$) with identical train, validation, and test partitions; LAAT keeps its frozen MIMIC-IV encoder and retrains only the label head. Adapted PLM-ICD (80.70\% strict) is reported in Supplementary Table S24; CAML was not adapted because it lacks a comparably defined frozen-encoder/head split. In-domain seven-category results (Table~\ref{tab:mimic3_baselines}) are not comparable to this decision space. The same comparison with the MedFound sensitivity backbone is in Supplementary Table S21, and zero-shot text-baseline transfer under an unrestricted seven-way decision is in Supplementary Table S23. No paired test was run between the MedGemma adapter and adapted LAAT, so their difference is
reported as a point estimate.
\item Medical accuracy equals strict accuracy here, since soft-matching adds no credit across the three coarse categories.
\item Label-efficiency (91.3\% at 5\% of target-training labels) and paired tests are in Supplementary Table S16. Macro AUROC/AUPRC are one-vs-rest averages.
\end{tablenotes}
\end{threeparttable}
\end{minipage}
\end{table}

The Combined adapter reached 92.51\% mapped accuracy at full supervision, exceeding adapted LAAT (85.54\%); label-efficiency results are in Supplementary Table S16.



\section{Discussion}

\subsection{Principal Findings}

This study shows that frozen LLM representations organize clinically useful information from structured and narrative EHR inputs in a form accessible to lightweight probes. Across MIMIC-IV, in-domain MIMIC-III, and cross-dataset transfer, the Combined pipeline achieved the best performance in every primary comparison, including when target labels for adaptation were limited.

Three observations support these conclusions. The Combined configuration improved most clearly over Structured-only probing, while its smaller margin over Unstructured-only indicates that discharge notes already contain much of the diagnostic signal. The same modality pattern reproduced across six frozen backbones spanning model families, training domains, and parameter scales, so the conclusion is framework-level rather than checkpoint-specific. And MIMIC-III supplied two complementary forms of evidence: in-domain replication against all three text baselines, and label-efficient reuse of frozen MIMIC-IV representations through a compact adapter under cross-dataset shift.

Because the task-specific component is only a linear probe, the framework's performance tracks the quality of the backbone: the two largest backbones achieved the highest terminal-layer Combined accuracies (Supplementary Table S12), and moving the entire analysis from an 8B to a 27B backbone required only re-extracting embeddings and retraining the probes on CPU, with the principal conclusions preserved. The framework therefore inherits improvements in medical LLMs at little engineering cost, whereas task-specific architectures require redesign and retraining to benefit from backbone progress.

\subsection{Comparison with Prior Work}

Automated ICD coding has been dominated by systems trained on discharge summaries, from convolutional (CAML) and recurrent (LAAT) label attention to pretrained transformers (PLM-ICD)~\cite{mullenbach2018explainable,vu2020label,huang2022plm}. That these three architectures performed within a narrow band in our budget-matched setting, with LAAT strongest, indicates that the reference point is a competitive one, and a linear probe on frozen representations sat at the top of it. Multimodal EHR studies have shown that structured variables can complement notes through late fusion, separate encoders, cross-modal pretraining, or specialized transformer architectures~\cite{choi2016doctor,choi2016retain,scheurwegs2016data,liu2022multimodal,lyu2023multimodal,merchant2024ensemble,thao2024medfuse}; our results show that a comparable multimodal gain is obtainable without a fusion architecture, by serializing the structured variables and encoding both modalities with one frozen LLM, and that the gain persists when the baselines receive the same serialized input. The approach builds on recent medical LLM and EHR foundation-model work~\cite{liu2025generalist,acharya2024clinical,redekop2025zero,kim2026medrep}.

\subsection{Clinical Implications}

Clinically, the framework is most useful as a representation analysis and adaptation tool rather than an autonomous ICD coding system. Because the LLM backbone remains frozen, the same extracted embeddings can support repeated analyses of layers, modalities, prompt formats, and local adaptation without repeatedly fine-tuning a large task-specific model, helping investigators determine whether structured values, clinical narratives, or their combination carry signal for a target phenotype. In the mapped MIMIC-III transfer experiment, 361 labeled target-training examples,
alongside the retained validation set, were sufficient for the Combined adapter to reach 91.3\% accuracy and retain an advantage over matched text-only adaptation, suggesting a practical route for representation realignment when local labels are limited, although broader institutional validation remains necessary.

\subsection{Limitations}

Several limitations should be considered. First, the prediction task was restricted to the 10 most prevalent primary ICD codes, consolidated into seven categories. This controlled setting improves interpretability for representation probing, but it does not capture the full multi-label and long-tail structure of production ICD coding. Second, both MIMIC-IV and MIMIC-III are retrospective critical-care datasets, so performance may not generalize to outpatient settings, non-ICU populations, or institutions with different documentation practices. The canonical partition is admission-level rather than grouped by patient, so repeated admissions from one patient could occur in different partitions. On MIMIC-III, a patient-grouped resplit lowered Combined terminal-layer medical accuracy only from 88.84\% to 87.85\% with the configuration ordering preserved, so patient overlap does not materially inflate the results. Third, template-based serialization compresses structured EHR trajectories into text and may lose clinically important temporal patterns; performance also varied with context budget. 
Fourth, medical accuracy is a study-specific relaxed matching metric rather than a standardized clinical endpoint and should therefore be interpreted alongside strict accuracy and per-class recall. 
Fifth, the six frozen decoder backbones span only the Llama and Gemma families; encoder-only, mixture-of-experts, and specialized EHR foundation models were not evaluated, and the scaling observation is descriptive, since the 27B and 70B backbones performed comparably on this near-saturated seven-category task. Sixth, in the mapped transfer analysis every label fraction retains the full validation set, zero-shot and adapted comparisons use different output spaces, and fractional adapter runs are single realizations. Seventh, the documented pretraining data of the primary backbone (MedGemma-27B) include MIMIC-CXR radiology reports but no MIMIC discharge summaries, so the notes used here are absent from its documented corpus. The MedFound sensitivity backbone was pretrained on MedCorpus, which includes MIMIC-III discharge notes, so its MIMIC-III sensitivity results may be advantaged; Meditron3 documents no MIMIC sources. Probe labels, splits, and evaluation data were never exposed to any backbone. Finally, the study does not include prospective validation or assessment of workflow integration with professional coders.

\section{Conclusion}
We presented a general multimodal probing framework that represents structured EHR variables and clinician-authored narratives within a shared frozen-LLM space and recovers primary diagnosis categories using lightweight linear probes. Across three input configurations, six frozen backbones spanning different model families and scales, two EHR datasets, and a cross-dataset adaptation setting, the Combined configuration consistently produced the strongest terminal-layer representations and matched or outperformed task-specific baselines in every comparison. Its advantage over both single-modality configurations indicates that the framework can exploit complementary diagnostic information from structured and narrative sources without requiring a bespoke multimodal fusion architecture. The resulting representations were also reusable across datasets: a compact adapter recovered strong performance on MIMIC-III with limited target-domain supervision while keeping both the source backbone and probe frozen. These findings position the framework as a practical methodology for identifying where diagnostic signal resides across EHR modalities, evaluating the incremental value of additional data sources, and adapting clinical representations when labeled data are scarce. Future work should extend the evaluation to full multi-label and long-tail ICD coding, broader model families, richer temporal representations, and prospective validation across multiple institutions.


\section*{Conflicts of Interest}
None declared.

\section*{Funding}
No external funding was received.

\section*{Data Availability}
MIMIC-IV v3.1 and MIMIC-III v1.4 are available through PhysioNet credentialed access and data-use agreements. Patient-level data cannot be redistributed. Analysis code, cohort-construction and serialization scripts, derived aggregate metrics, and the exact identifiers of all evaluated model checkpoints will be released publicly upon acceptance.

\section*{Author contributions}
Chengyuan Liu (Conceptualization, Data curation, Methodology, Software, Visualization, Writing---original draft, Writing---review \& editing), Xinyue Zhang (Data curation, Methodology, Visualization), Yao Li (Methodology, Project administration, Supervision, Writing---review \& editing), and Guanting Chen (Conceptualization, Formal analysis, Investigation, Methodology, Project administration, Supervision, Writing---original draft, Writing---review \& editing).

\section*{Ethics approval and patient consent}
Since this study only used publicly available datasets, informed consent and institutional review board (IRB) approval were not required.

\bibliographystyle{jamia_submission}
\bibliography{reference_jamia}

\begin{thebibliography}{34}
\providecommand{\natexlab}[1]{#1}
\providecommand{\url}[1]{\texttt{#1}}
\expandafter\ifx\csname urlstyle\endcsname\relax
  \providecommand{\doi}[1]{doi: #1}\else
  \providecommand{\doi}{doi: \begingroup \urlstyle{rm}\Url}\fi

\bibitem[{World Health Organization}(1992)]{who1992international}
{World Health Organization}.
\newblock International classification of diseases.
\newblock \emph{WHO [Internet]} 1992:53.

\bibitem[Adiba et~al.(2026)Adiba, Danduri, Piya, Abbasi, Gupta, and Beheshti]{adiba2026multimodal}
Adiba FI, Danduri V, Piya FL, et~al.
\newblock A multimodal data processing pipeline for {MIMIC-IV} dataset.
\newblock \emph{arXiv [Preprint]} 2026.
\newblock arXiv:2601.11606.

\bibitem[Dou et~al.(2024)Dou, Li, Zheng, Yao, Liu, Zomaya, and Peng]{dou2024quick}
Dou Y, Li W, Zheng Y, et~al.
\newblock {Quick-MIMIC}: A multimodal data extraction pipeline for {MIMIC} with parallelization.
\newblock \emph{Big Data Mining and Analytics} 2024;7(4):1333--1346.

\bibitem[Xu et~al.(2019)Xu, Lam, Pang, Gao, Band, Mathur, Papay, Khanna, Cywinski, Maheshwari, et~al.]{xu2019multimodal}
Xu K, Lam M, Pang J, et~al.
\newblock Multimodal machine learning for automated {ICD} coding.
\newblock In: Machine Learning for Healthcare Conference, PMLR, 2019:197--215.

\bibitem[Liu et~al.(2022{\natexlab{a}})Liu, Liu, Wen, Zhao, Xia, and Yuan]{liu2022treeman}
Liu Z, Liu X, Wen Y, et~al.
\newblock {TreeMAN}: tree-enhanced multimodal attention network for {ICD} coding.
\newblock In: Proceedings of the 29th International Conference on Computational Linguistics, 2022{\natexlab{a}}:3054--3063.

\bibitem[Liu et~al.(2022{\natexlab{b}})Liu, Wang, Hou, Li, Wang, Xu, Xiang, and Tang]{liu2022multimodal}
Liu S, Wang X, Hou Y, et~al.
\newblock Multimodal data matters: Language model pre-training over structured and unstructured electronic health records.
\newblock \emph{IEEE J Biomed Health Inform} 2022{\natexlab{b}};27(1):504--514.

\bibitem[Merchant et~al.(2024)Merchant, Shenoy, Lanka, and Kamath]{merchant2024ensemble}
Merchant AM, Shenoy N, Lanka S, et~al.
\newblock Ensemble neural models for {ICD} code prediction using unstructured and structured healthcare data.
\newblock \emph{Heliyon} 2024;10(17).

\bibitem[Thao et~al.(2024)Thao, Dao, Wu, Wang, Liu, Ding, Restrepo, Liu, Hung, and Peng]{thao2024medfuse}
Thao PNM, Dao CT, Wu C, et~al.
\newblock {MedFuse}: Multimodal {EHR} data fusion with masked lab-test modeling and large language models.
\newblock In: Proceedings of the 33rd ACM International Conference on Information and Knowledge Management, 2024:3974--3978.

\bibitem[Ruan et~al.(2025)Ruan, Lan, Tan, Abdullah, and Feng]{ruan2025ptransformer}
Ruan Y, Lan X, Tan DJ, et~al.
\newblock {PTransformer}: a prompt-based multimodal transformer architecture for medical tabular data.
\newblock In: International Conference on Artificial Intelligence in Medicine, Springer, 2025:375--385.

\bibitem[Lee et~al.(2025)Lee, Jain, Chen, Ono, Biswas, Rudas, Fang, and Chiang]{lee2025clinical}
Lee SA, Jain S, Chen A, et~al.
\newblock Clinical decision support using pseudo-notes from multiple streams of {EHR} data.
\newblock \emph{NPJ Digit Med} 2025;8(1):394.

\bibitem[Zhu et~al.(2026)Zhu, Gao, Wang, Liao, Zheng, Liang, Bernabeu, Wang, Yu, Pan, et~al.]{zhu2026clinicrealm}
Zhu Y, Gao J, Wang Z, et~al.
\newblock {ClinicRealm}: Re-evaluating large language models with conventional machine learning for non-generative clinical prediction tasks.
\newblock \emph{NPJ Digit Med} 2026.

\bibitem[Hegselmann et~al.(2026)Hegselmann, von Arnim, Rheude, Kronenberg, Sontag, Hindricks, Eils, and Wild]{hegselmann2025large}
Hegselmann S, von Arnim G, Rheude T, et~al.
\newblock Large language models are powerful electronic health record encoders.
\newblock \emph{NPJ Digit Med} 2026.
\newblock \doi{10.1038/s41746-026-02915-9}.

\bibitem[Engelke et~al.(2025)Engelke, Baldini, Kleesiek, Nensa, and Dada]{engelke2025fhir}
Engelke M, Baldini G, Kleesiek J, et~al.
\newblock {FHIR}-former: enhancing clinical predictions through fast healthcare interoperability resources and large language models.
\newblock \emph{J Am Med Inform Assoc} 2025;32(12):1793--1801.

\bibitem[Johnson et~al.(2024)Johnson, Bulgarelli, Pollard, Gow, Moody, Horng, Celi, and Mark]{mimiciv31}
[dataset] Johnson A, Bulgarelli L, Pollard T, et~al.
\newblock {MIMIC-IV} (version 3.1).
\newblock PhysioNet 2024.
\newblock \doi{10.13026/kpb9-mt58}.

\bibitem[Johnson et~al.(2016)Johnson, Pollard, and Mark]{mimiciii14}
[dataset] Johnson A, Pollard T, Mark R.
\newblock {MIMIC-III Clinical Database} (version 1.4).
\newblock PhysioNet 2016.
\newblock \doi{10.13026/C2XW26}.

\bibitem[Rao et~al.(2025)Rao, O’Donoghue, Ruel, Rab, Tamis-Holland, Alexander, Baber, Baker, Cohen, Cruz-Ruiz, et~al.]{rao20252025}
Rao SV, O’Donoghue ML, Ruel M, et~al.
\newblock 2025 {ACC/AHA/ACEP/NAEMSP/SCAI} guideline for the management of patients with acute coronary syndromes: a report of the {American College of Cardiology/American Heart Association Joint Committee on Clinical Practice Guidelines}.
\newblock \emph{J Am Coll Cardiol} 2025;85(22):2135--2237.

\bibitem[Rangaswami et~al.(2019)Rangaswami, Bhalla, Blair, Chang, Costa, Lentine, Lerma, Mezue, Molitch, Mullens, et~al.]{rangaswami2019cardiorenal}
Rangaswami J, Bhalla V, Blair JE, et~al.
\newblock Cardiorenal syndrome: classification, pathophysiology, diagnosis, and treatment strategies: a scientific statement from the {American Heart Association}.
\newblock \emph{Circulation} 2019;139(16):e840--e878.

\bibitem[Zhang et~al.(2021)Zhang, Zhang, Xu, Wang, Ren, Han, Lyu, and Yin]{zhang2021construction}
Zhang L, Zhang F, Xu F, et~al.
\newblock Construction and evaluation of a sepsis risk prediction model for urinary tract infection.
\newblock \emph{Front Med (Lausanne)} 2021;8:671184.

\bibitem[Alain and Bengio(2016)]{alain2016understanding}
Alain G, Bengio Y.
\newblock Understanding intermediate layers using linear classifier probes.
\newblock \emph{arXiv [Preprint]} 2016.
\newblock arXiv:1610.01644.

\bibitem[Hewitt and Liang(2019)]{hewitt2019designing}
Hewitt J, Liang P.
\newblock Designing and interpreting probes with control tasks.
\newblock In: Proceedings of the 2019 Conference on Empirical Methods in Natural Language Processing and the 9th International Joint Conference on Natural Language Processing (EMNLP-IJCNLP), 2019:2733--2743.

\bibitem[Chen and Guestrin(2016)]{chen2016xgboost}
Chen T, Guestrin C.
\newblock {XGBoost}: A scalable tree boosting system.
\newblock In: Proceedings of the 22nd ACM SIGKDD International Conference on Knowledge Discovery and Data Mining, 2016:785--794.

\bibitem[Mullenbach et~al.(2018)Mullenbach, Wiegreffe, Duke, Sun, and Eisenstein]{mullenbach2018explainable}
Mullenbach J, Wiegreffe S, Duke J, et~al.
\newblock Explainable prediction of medical codes from clinical text.
\newblock In: Proceedings of the 2018 Conference of the North American Chapter of the Association for Computational Linguistics: Human Language Technologies, Volume 1 (Long Papers), 2018:1101--1111.

\bibitem[Vu et~al.(2020)Vu, Nguyen, and Nguyen]{vu2020label}
Vu T, Nguyen DQ, Nguyen A.
\newblock A label attention model for {ICD} coding from clinical text.
\newblock In: Proceedings of the Twenty-Ninth International Joint Conference on Artificial Intelligence (IJCAI-20), 2020:3335--3341.
\newblock \doi{10.24963/ijcai.2020/461}.

\bibitem[Huang et~al.(2022)Huang, Tsai, and Chen]{huang2022plm}
Huang CW, Tsai SC, Chen YN.
\newblock {PLM-ICD}: Automatic {ICD} coding with pretrained language models.
\newblock In: Proceedings of the 4th Clinical Natural Language Processing Workshop, Seattle, WA, Association for Computational Linguistics, 2022:10--20.
\newblock \doi{10.18653/v1/2022.clinicalnlp-1.2}.

\bibitem[Rajkomar et~al.(2018)Rajkomar, Oren, Chen, Dai, Hajaj, Hardt, Liu, Liu, Marcus, Sun, et~al.]{rajkomar2018scalable}
Rajkomar A, Oren E, Chen K, et~al.
\newblock Scalable and accurate deep learning with electronic health records.
\newblock \emph{NPJ Digit Med} 2018;1(1):18.

\bibitem[Harutyunyan et~al.(2019)Harutyunyan, Khachatrian, Kale, Ver~Steeg, and Galstyan]{harutyunyan2019multitask}
Harutyunyan H, Khachatrian H, Kale DC, et~al.
\newblock Multitask learning and benchmarking with clinical time series data.
\newblock \emph{Sci Data} 2019;6(1):96.

\bibitem[Lyu et~al.(2023)Lyu, Dong, Wong, Zheng, Abell-Hart, Wang, and Chen]{lyu2023multimodal}
Lyu W, Dong X, Wong R, et~al.
\newblock A multimodal transformer: Fusing clinical notes with structured {EHR} data for interpretable in-hospital mortality prediction.
\newblock In: AMIA Annu Symp Proc, volume 2022, 2023:719.

\bibitem[Choi et~al.(2016{\natexlab{a}})Choi, Bahadori, Schuetz, Stewart, and Sun]{choi2016doctor}
Choi E, Bahadori MT, Schuetz A, et~al.
\newblock Doctor {AI}: Predicting clinical events via recurrent neural networks.
\newblock In: Machine Learning for Healthcare Conference, PMLR, 2016{\natexlab{a}}:301--318.

\bibitem[Choi et~al.(2016{\natexlab{b}})Choi, Bahadori, Sun, Kulas, Schuetz, and Stewart]{choi2016retain}
Choi E, Bahadori MT, Sun J, et~al.
\newblock {RETAIN}: An interpretable predictive model for healthcare using reverse time attention mechanism.
\newblock \emph{Advances in neural information processing systems} 2016{\natexlab{b}};29.

\bibitem[Scheurwegs et~al.(2016)Scheurwegs, Luyckx, Luyten, Daelemans, and Van~den Bulcke]{scheurwegs2016data}
Scheurwegs E, Luyckx K, Luyten L, et~al.
\newblock Data integration of structured and unstructured sources for assigning clinical codes to patient stays.
\newblock \emph{J Am Med Inform Assoc} 2016;23(e1):e11--e19.

\bibitem[Liu et~al.(2025)Liu, Liu, Yang, Jiang, Cui, Zhang, Wang, Tao, Sun, Song, et~al.]{liu2025generalist}
Liu X, Liu H, Yang G, et~al.
\newblock A generalist medical language model for disease diagnosis assistance.
\newblock \emph{Nat Med} 2025;31(3):932--942.

\bibitem[Acharya et~al.(2024)Acharya, Shrestha, Chen, Conte, Avramovic, Sikdar, Anastasopoulos, and Das]{acharya2024clinical}
Acharya A, Shrestha S, Chen A, et~al.
\newblock Clinical risk prediction using language models: benefits and considerations.
\newblock \emph{J Am Med Inform Assoc} 2024;31(9):1856--1864.

\bibitem[Redekop et~al.(2025)Redekop, Wang, Kulkarni, Pleasure, Chin, Hassanzadeh, Hill, Emami, Speier, and Arnold]{redekop2025zero}
Redekop E, Wang Z, Kulkarni R, et~al.
\newblock Zero-shot medical event prediction using a generative pretrained transformer on electronic health records.
\newblock \emph{J Am Med Inform Assoc} 2025;32(12):1833--1842.

\bibitem[Kim et~al.(2026)Kim, Lee, Kim, and Kim]{kim2026medrep}
Kim J, Lee N, Kim J, et~al.
\newblock Medrep: medical concept representations for general electronic health record foundation models.
\newblock \emph{J Am Med Inform Assoc} 2026:ocag032.

\end{thebibliography}

\clearpage
\appendix
\raggedbottom
\setcounter{table}{0}
\renewcommand{\tablename}{Supplementary Table}
\renewcommand{\thetable}{S\arabic{table}}
\setcounter{section}{0}
\renewcommand{\thesection}{}
\section*{\LARGE Supplementary Materials}
\addcontentsline{toc}{section}{Supplementary Materials}
\section*{Supplementary Methods}

\subsection*{Detailed Cohort Construction}

We used MIMIC-IV v3.1 as the primary dataset and MIMIC-III v1.4 for two analyses. First, we trained and evaluated linear probes on separate MIMIC-III splits to assess performance under its documentation style and ICD-9 coding system. Second, we adapted the MIMIC-IV probe to MIMIC-III using a compact adapter and compared it with text-baseline adaptation. The LLM backbone remained frozen in both analyses. Both databases are de-identified critical-care EHR resources accessed through PhysioNet credentialing and data-use agreements.

The MIMIC-IV cohort was constructed at the hospital-admission level. For each admission, we identified the primary diagnosis from records with primary-code rank equal to 1 and retained one primary diagnosis per \texttt{hadm\_id}. We then ranked primary ICD-10 codes by frequency and selected admissions belonging to the 10 most prevalent primary codes in the resulting cohort. This yielded 13,645 admissions for the main MIMIC-IV analysis.

MIMIC-III was processed analogously using the 10 most prevalent ICD-9 primary diagnosis codes. The same admission-level framing, label consolidation logic, and train/validation/test workflow were used to maintain comparable evaluation conditions across datasets. Cross-dataset adaptation used the separately defined mapped three-category task described below and in Supplementary Tables S8, S16, and S23--S24.

For MIMIC-IV model development, we used a canonical admission-level stratified split of 70\%/15\%/15\%, corresponding to 9,556 training admissions, 2,042 validation admissions, and 2,047 test admissions. Cohort characteristics, category prevalence, and split sizes are summarized in Supplementary Table S1.

\begingroup
\small
\setlength{\tabcolsep}{3pt}
\setlength{\suppwidth}{\dimexpr\linewidth-4\tabcolsep\relax}
\renewcommand{\arraystretch}{1.05}
\begin{longtable}{@{}L{0.510000\suppwidth}L{0.250000\suppwidth}R{0.240000\suppwidth}@{}}
\caption{Cohort characteristics and ICD-10 diagnosis categories for the MIMIC-IV prediction cohort.}\label{tab:supp_cohort_characteristics}\\
\toprule
\textbf{Characteristic or category} & \textbf{ICD-10 code(s)} & \textbf{Value} \\
\midrule
\endfirsthead
\multicolumn{3}{@{}l}{\small\textit{Supplementary Table S1 (continued)}}\\
\toprule
\textbf{Characteristic or category} & \textbf{ICD-10 code(s)} & \textbf{Value} \\
\midrule
\endhead
\midrule
\multicolumn{3}{r@{}}{\footnotesize Continued on next page}\\
\endfoot
\bottomrule
\endlastfoot
\multicolumn{3}{l}{\textbf{Cohort characteristics}} \\
Total admissions & -- & 13,645 \\
Age, mean $\pm$ SD, years & -- & 64.0 $\pm$ 16.4 \\
Male sex & -- & 7,212 (52.9\%) \\
Female sex & -- & 6,433 (47.1\%) \\
\midrule
\multicolumn{3}{l}{\textbf{Diagnosis categories}} \\
Infection Related & A419; N390 & 4,326 (31.7\%) \\
Hypertensive Heart Failure & I110; I130 & 2,902 (21.3\%) \\
Cardiovascular Event & R079; I214 & 2,241 (16.4\%) \\
Chemotherapy Encounter & Z5111 & 2,163 (15.9\%) \\
Acute Kidney Failure & N179 & 1,596 (11.7\%) \\
Major Depression & F329 & 368 (2.7\%) \\
Alcohol Intoxication & F10129 & 49 (0.4\%) \\
\midrule
\multicolumn{3}{l}{\textbf{Data split}} \\
Training & -- & 9,556 (70.0\%) \\
Validation & -- & 2,042 (15.0\%) \\
Test & -- & 2,047 (15.0\%) \\
\end{longtable}
\nopagebreak[3]
{\footnotesize\raggedright
\noindent ICD-10 code descriptions and per-code case counts are provided in Supplementary Table S3.\par}

\endgroup

\subsection*{Structured Data Serialization Details}

For each admission $i$, structured features were organized as demographics, medications, laboratory results, and summary statistics. We converted these features into a fixed clinical narrative template to standardize tokenized input for the frozen language-model backbone, preserve cross-patient consistency, and keep prompt length stable across samples.

Template truncation rules prioritized clinically salient variables while controlling sequence length: medications were limited to the 20 most recent prescriptions and laboratory content to 25 tests with abnormal values prioritized. The resulting structured narratives averaged 1,139 characters.

The structured template emits the following fields in fixed order, joined by the delimiter \texttt{ | }, and omits a field when the underlying table contributes no rows for that admission:

\begingroup\ttfamily\raggedright\small Patient: \textnormal{$\langle$age$\rangle$}-year-old \textnormal{$\langle$sex$\rangle$} | Current medications: \textnormal{$\langle$up to 20 most recent prescriptions$\rangle$} | Laboratory results: \textnormal{$\langle$up to 25 tests, abnormal values first$\rangle$} | Total prescriptions: \textnormal{$\langle$count$\rangle$} | Laboratory summary: \textnormal{$\langle$counts of normal and abnormal results$\rangle$}\endgroup

In the MIMIC-IV cohort this full five-field form occurs for the large majority of admissions; admissions with no prescription records reduce to the Patient, Laboratory results and Laboratory summary fields, and admissions with no laboratory records reduce to the Patient, Current medications and Total prescriptions fields. Field labels are constant across admissions so that tokenization and prompt length remain comparable.

\subsection*{Unstructured Section Extraction and Leakage Control}

For discharge notes, we removed diagnosis-bearing discharge-tail content before constructing model input. Truncation started at the earliest occurrence of markers including ``Discharge Disposition,'' ``Discharge Diagnosis,'' ``PRIMARY DIAGNOSIS,'' ``Discharge Condition,'' ``Discharge Instructions,'' and ``Followup Instructions.''

After leakage pruning, we extracted and concatenated five primary sections: Chief Complaint, History of Present Illness, Physical Exam, Brief Hospital Course, and Pertinent Results. This produced leakage-mitigated unstructured narratives averaging approximately 2,000 characters.

Sections are identified by their heading text at the start of a line and delimited by the next recognized heading or by the truncation markers listed above, whichever comes first. Only the five headings above are treated as section boundaries; sub-headings that appear inside them, such as the system-by-system entries of the physical examination or the imaging impression within the pertinent results, are retained as part of their parent section rather than split out. Extracted sections are joined with the same \texttt{ | } delimiter used by the structured template, so Structured-only, Unstructured-only and Combined inputs share one surface format.

\subsection*{Context-Budget Truncation}

When an assembled prompt exceeds the 3,000-character budget it is reduced by section-priority selection rather than by truncating the tail. The delimiter is detected automatically, \texttt{ | } for MIMIC-IV-style inputs and a newline for MIMIC-III-style inputs. Sections are then considered in the fixed priority order Chief Complaint, History of Present Illness, Physical Exam, Current medications, Medications, Assessment and Plan, Labs, Laboratory results, and each is kept in full if it fits in the remaining budget. Selection stops once fewer than 100 characters remain. A section whose terminating delimiter cannot be located is capped at 400 characters. If no priority section is matched at all, the prompt falls back to head truncation at 3,000 characters. Because selection is by whole sections, the retained text is always a subset of complete sections rather than a fragment.

\subsection*{Example Leakage-Pruned Narrative Format}

The example below illustrates the final section-structured unstructured input format after discharge-tail leakage pruning and section extraction.

\begin{tcolorbox}[colback=gray!10, colframe=gray!70, title=Example]
Chief Complaint: shortness of breath \\
History of Present Illness: Patient is [age] years old with a history of coronary artery disease... \\
Physical Exam: Pleasant female appearing younger than her stated age... \\
Brief Hospital Course: [clinical course details] \\
Pertinent Results: [lab values and imaging].
\end{tcolorbox}

\subsection*{Prompt Construction Details}

Three configuration-specific prompts were used:
\[
P_{\text{str}}(X_i)=T_{\text{str}}(X_i^{\text{str}}), \quad
P_{\text{txt}}(X_i)=T_{\text{txt}}(X_i^{\text{txt}}), \quad
P_{\text{combined}}(X_i)=P_{\text{str}}(X_i)\oplus P_{\text{txt}}(X_i),
\]
where $\oplus$ denotes text concatenation with a delimiter. Each prompt was appended with the fixed query ``What's the primary diagnosis?'' to maintain consistent prediction context across configurations.

\subsection*{Backbone Robustness}

The primary analysis used frozen MedGemma-27B (62 layers; representation width 5,376), with probes at layers 2, 16, 31, 47, and 62. The MedFound-Llama3-8B-finetuned sensitivity analysis used layers 1, 8, 16, 24, and 32 (representation width 4,096). Terminal-layer robustness experiments also evaluated Meta-Llama-3-8B (Layer 32; width 4,096), Llama-3.1-8B (Layer 32; width 4,096), Meditron3-8B (Layer 32; width 4,096), and Llama-3.1-70B (Layer 80; width 8,192). Across all six frozen backbones, cohort, canonical split, input construction, mean-pooling strategy, and linear-probe specification were held fixed; each model used its own tokenizer.

The evaluated checkpoints were as follows, all loaded in bfloat16 with inference-only settings:

\begingroup\ttfamily\raggedright\small
google/medgemma-27b-it \textnormal{(primary backbone)}\\
medicalai/MedFound-Llama3-8B-finetuned\\
meta-llama/Meta-Llama-3-8B\\
meta-llama/Llama-3.1-8B\\
meta-llama/Llama-3.1-70B\\
EPFLiGHT/Meditron3-8B\\
\endgroup

Repository identifiers rather than pinned revisions are reported, since the runs did not record commit hashes; exact identifiers will be released with the analysis code.

\subsection*{Probe Training and Optimization}

All linear probes were trained with Adam using a learning rate of $10^{-4}$, weight decay $10^{-5}$, batch size 16, and a maximum of 50 epochs with early stopping on validation accuracy at a patience of 10 epochs and gradient-norm clipping at 1.0. Probes were selected by validation accuracy and evaluated once on the held-out test admissions. Optimization settings were identical for the MIMIC-IV and MIMIC-III analyses and across all backbones, layers, and input configurations, so differences between configurations reflect the frozen representations rather than probe tuning. Inputs were capped at 3,000 characters under the section-priority truncation described above, and the random seed for cohort splitting was fixed at 42.

\subsection*{Probe and Input Sensitivity}

Probe stability was evaluated on cached MedFound Combined Layer-32 embeddings using learning rates $5\times10^{-5}$, $10^{-4}$, and $2\times10^{-4}$ and random seeds 42, 7, and 123. All nine probes used the same canonical admission split and otherwise identical optimization settings. Input sensitivity compared the 3,000-character primary Combined prompt with a 2,000-character version and with a prompt in which modality order was reversed.

\subsection*{Evaluation Rule and Statistical Procedure Details}

Strict accuracy was defined as exact seven-category top-1 agreement with reference labels. For the seven-category MIMIC-IV task, medical accuracy credited exact matches and two predefined cross-category pairs: Cardiovascular Event with Hypertensive Heart Failure, and Acute Kidney Failure with Hypertensive Heart Failure. All other cross-category predictions were errors. The same rule was applied to every model in this label space. The mapped three-category transfer task used strict accuracy; medical accuracy was identical because no additional pair was credited.

For the seven-category MIMIC-III (ICD-9) label space, the same soft-matching implementation credits six symmetric cross-category pairs: Congestive Heart Failure with Coronary Artery Disease; Congestive Heart Failure with Respiratory Failure; Congestive Heart Failure with Valvular Heart Disease; Coronary Artery Disease with Valvular Heart Disease; Pneumonia with Respiratory Failure; and Respiratory Failure with Valvular Heart Disease. Unspecified Septicemia and Intracerebral Hemorrhage receive no cross-category credit. Because six of the 21 possible pairs are credited on MIMIC-III against two of 21 on MIMIC-IV, the MIMIC-III rule is substantially more permissive, and the gap between strict and medical accuracy is correspondingly wider there (74.18\% versus 88.84\% for the primary Combined probe, against 88.86\% versus 91.79\% on MIMIC-IV). Medical accuracies should therefore be compared within a dataset and not across the two datasets, and strict accuracy is the primary metric in both.

Paired model comparisons on the same test admissions used exact McNemar tests for strict top-1 accuracy and paired bootstrap resampling for differences in accuracy, macro accuracy, AUROC, and AUPRC. Bootstrap confidence intervals were computed by resampling admissions with replacement and recomputing both models on the same resampled cohort. Multiple planned pairwise tests within each analysis family were adjusted with the Benjamini-Hochberg false discovery rate procedure. The number of bootstrap resamples was fixed within each analysis family but differs between families: 10,000 for the probing-family comparisons (Supplementary Table S5), 5,000 for the MIMIC-IV text-baseline comparisons (Supplementary Tables S17 and S18), and 2,000 for the MIMIC-III text-baseline and adapter-transfer comparisons (Supplementary Tables S20 and S24). Resampling used a fixed seed so that intervals are reproducible. The diagnosis-keyword masking ablation is the one exception to the metric-to-test assignment above: because it compares two runs of the same model over identical admissions and an identical label space, differences in both strict and medical accuracy were assessed with exact McNemar tests on the paired predictions.

Supplementary Table S2 specifies, for each model family, the training label space, the number of model outputs, and the label space on which every reported metric is computed.

\begingroup
\small
\setlength{\tabcolsep}{3pt}
\setlength{\suppwidth}{\dimexpr\linewidth-6\tabcolsep\relax}
\renewcommand{\arraystretch}{1.05}
\begin{longtable}{@{}L{0.300000\suppwidth}C{0.200000\suppwidth}C{0.110000\suppwidth}L{0.390000\suppwidth}@{}}
\caption{Label and evaluation contract for all MIMIC-IV models.}\label{tab:supp_label_contract}\\
\toprule
\textbf{Model} & \textbf{Training labels} & \textbf{Outputs} & \textbf{Top-1 prediction} \\
\midrule
\endfirsthead
\multicolumn{4}{@{}l}{\small\textit{Supplementary Table S2 (continued)}}\\
\toprule
\textbf{Model} & \textbf{Training labels} & \textbf{Outputs} & \textbf{Top-1 prediction} \\
\midrule
\endhead
\midrule
\multicolumn{4}{r@{}}{\footnotesize Continued on next page}\\
\endfoot
\bottomrule
\endlastfoot
Linear probes (all configurations) & 7 merged categories & 7 & argmax over 7 categories \\
XGBoost & 7 merged categories & 7 & argmax over 7 categories \\
CAML / LAAT / PLM-ICD & 10 raw ICD-10 codes & 10 & scores summed within the 7 groups; merged argmax \\
\end{longtable}
\nopagebreak[3]
{\footnotesize\raggedright
\noindent All models are evaluated on the same canonical test admissions ($n=2{,}047$) in the same seven-category space. Strict accuracy is exact seven-category top-1 match. Medical accuracy equals strict accuracy plus the two credited category pairs described in the main text. Macro AUROC and AUPRC are one-vs-rest over the seven categories, using group-summed scores for the text baselines. The MIMIC-III analyses also consolidate 10 raw ICD-9 codes into seven evaluation categories; the mapped transfer task uses three categories.\par}

\endgroup

\subsection*{Structured Baseline Preprocessing}

For the XGBoost structured baseline, the input feature matrix was derived from the MIMIC-IV \texttt{labevents} and \texttt{prescriptions} tables. We selected the 50 most frequent laboratory tests and vital signs to focus on high-yield clinical signals and reduce sparsity. Because EHR measurements are irregularly sampled, longitudinal values were converted into fixed-length vectors using summary statistics, including mean, maximum, minimum, and standard deviation. Missing values were retained as \texttt{NaN} rather than mean-imputed so that XGBoost's native sparsity-aware split finding could learn default branch directions and use informative missingness when predictive.

\subsection*{CAML and LAAT Adaptation}

We adapted the pinned official CAML implementation (commit \texttt{44a4745}) and official LAAT implementation (commit \texttt{cd5c0ec}) to the same PLM-ICD-matched text setting. Both consumed the canonical admission split, discharge-tail-pruned notes capped at approximately 3,000 characters, and a shared 100-dimensional CBOW Word2Vec model trained only on training notes. CAML retained its convolutional encoder (50 filters, kernel width 10) and label-specific attention; LAAT retained its one-layer bidirectional LSTM (512 hidden units per direction) and projected label attention. Both optimized binary cross-entropy over one-hot targets for the 10 raw primary ICD-10 codes. At evaluation, sigmoid scores were summed within the same seven predefined groups used for PLM-ICD, and the merged argmax defined the top-1 prediction. We ran seeds 42, 7, and 123; seed 42 was pre-specified for paired model comparisons. These are task adaptations of the published architectures, not reproductions of their original full-code MIMIC-III benchmarks.

For the MIMIC-III in-domain comparison, LAAT used the same official architecture, seeds, and train-only Word2Vec procedure on the 10 primary ICD-9 codes, followed by the same seven-category grouping as the probing analysis. Inputs were the section-structured Unstructured contexts used for MedFound extraction and the exact Combined-probe split. Because 1,056 admissions had no extracted unstructured context, they were retained with a fixed \texttt{<EMPTY>} placeholder rather than removed.

\subsection*{Cross-Dataset Adapter Specification}

In the bottleneck adapter $\hat{h}_i = \mathrm{LayerNorm}(h_i + W_{\mathrm{up}}\,\mathrm{ReLU}(W_{\mathrm{down}} h_i))$, $W_{\mathrm{up}}$ is zero-initialized, so at initialization the adapter output reduces to $\mathrm{LayerNorm}(h_i)$ rather than an exact identity map; training then learns a residual correction on top of this normalized representation. Only the adapter parameters are optimized; the source backbone and the MIMIC-IV probe, including its decision boundaries, remain fixed.

\subsection*{Cross-Dataset Text-Baseline Transfer Controls}

As Unstructured-only controls for the mapped MIMIC-III transfer task, we evaluated frozen MIMIC-IV PLM-ICD and LAAT encoders under two protocols. The mapped cohort retained the five mappable MIMIC-III source diagnoses and three MIMIC-IV target categories (10,321 admissions; fixed split 7,228/1,544/1,549), the same leakage-pruned Unstructured notes, 3,000-character budget, and \texttt{<EMPTY>} policy. Zero-shot transfer applied the frozen source models to all mapped admissions, mapped raw ICD-10 scores to the seven MIMIC-IV groups, and scored predictions without restricting the decision space to the three valid target categories; these results are reported only as a lower-bound panel (Supplementary Table S23) and are not plotted as a continuous learning curve into adapted three-class accuracy. Target-supervised adaptation froze each source encoder, replaced only the label-specific head (PLM-ICD: 594K parameters; LAAT: 4.6K parameters), and trained the heads on nested subsets containing 5\%, 10\%, 25\%, 50\%, or 100\% of the training admissions while always retaining the full 1,544-label validation set; therefore the 5\% condition used 361 training plus 1,544 validation labels (1,905 labeled target admissions). Seeds 42, 7, and 123 were used for adapted heads, with seed 42 reserved for paired comparisons against saved Unstructured and Combined bottleneck-adapter predictions at 100\% (Supplementary Table S24). CAML was omitted because its convolution and label-specific attention do not yield a comparably defined frozen-encoder/head split.

\subsection*{Excluded Multimodal Comparators}

Beyond the per-modality baselines, the prior approach closest to ours jointly models structured and unstructured signals for ICD prediction (Xu et al., 2019), but it cannot serve as a controlled comparator: no runnable implementation is publicly available, and the specific subset of laboratory and medication variables it retains from the thousands present in MIMIC is not fully specified, so its preprocessing cannot be reproduced under our matched split and inputs. To the best of our knowledge, there is likewise no established pipeline that predicts primary ICD categories from structured data alone; widely used structured-EHR models instead target different endpoints---such as in-hospital mortality, readmission, or length-of-stay prediction---under a longitudinal, patient-level framing (Rajkomar et al., 2018; Harutyunyan et al., 2019) rather than our admission-level diagnosis task, which is why we adopt a tuned XGBoost as the conventional structured reference. More recent multimodal fusion architectures, including dedicated cross-modal encoders and ensembles (MedFuse, Thao et al., 2024; Lyu et al., 2023; Merchant et al., 2024), are built around fusion mechanisms, prediction targets, or training regimes we do not adopt: for example, Lyu et al.\ target in-hospital mortality, and MedFuse couples masked lab-test modeling with a large language model under its own prediction setup. Reproducing any of these under our matched split, inputs, and seven-category primary-diagnosis space would require substantial re-engineering, so we treat them as complementary directions rather than head-to-head comparators.

\ReserveSpace{270pt}
\subsection*{MIMIC-IV ICD-10 Diagnosis Grouping}

Supplementary Table S3 reports the full ICD-10 diagnosis grouping used to define the seven-category MIMIC-IV prediction task. Supplementary Table S1 reports the same categories in summarized form alongside the cohort characteristics.

\begingroup
\small
\setlength{\tabcolsep}{3pt}
\setlength{\suppwidth}{\dimexpr\linewidth-4\tabcolsep\relax}
\renewcommand{\arraystretch}{1.05}
\begin{longtable}{@{}L{0.290000\suppwidth}L{0.610000\suppwidth}R{0.100000\suppwidth}@{}}
\caption{ICD-10 diagnosis grouping based on clinical relationships. Case counts are reported in the final column.}\label{tab:supp_icd10_grouping}\\
\toprule
\textbf{Grouped Category} & \textbf{ICD-10 Codes and Descriptions} & \textbf{Cases} \\
\midrule
\endfirsthead
\multicolumn{3}{@{}l}{\small\textit{Supplementary Table S3 (continued)}}\\
\toprule
\textbf{Grouped Category} & \textbf{ICD-10 Codes and Descriptions} & \textbf{Cases} \\
\midrule
\endhead
\midrule
\multicolumn{3}{r@{}}{\footnotesize Continued on next page}\\
\endfoot
\bottomrule
\endlastfoot
Cardiovascular Event & R079: Chest pain, unspecified & 253 \\
                    & I214: Non-ST elevation (NSTEMI) myocardial infarction & 1,988 \\
\midrule
Hypertensive Heart Failure & I110: Hypertensive heart disease with heart failure & 1,169 \\
                           & I130: Hypertensive heart and CKD with heart failure & 1,733 \\
\midrule
Infection Related & A419: Sepsis, unspecified organism & 3,060 \\
                  & N390: Urinary tract infection, site not specified & 1,266 \\
\midrule
Acute Kidney Failure & N179: Acute kidney failure, unspecified & 1,596 \\
Alcohol Intoxication & F10129: Alcohol abuse with intoxication & 49 \\
Chemotherapy Encounter & Z5111: Encounter for antineoplastic chemotherapy & 2,163 \\
Major Depression & F329: Major depressive disorder, single episode & 368 \\
\end{longtable}

\endgroup

\section*{Supplementary Results}

The tables below follow the main text in treating MedGemma-27B as the primary backbone and MedFound-Llama3-8B as a sensitivity analysis. Supplementary Tables S4, S5, S6, S9, S11, S15, S16, S17, S18, S19, S20, S21, and S22 carry the primary MedGemma-27B results; where a MedFound counterpart exists it is retained in the same table under a labelled sensitivity heading, or as a separate table. Supplementary Tables S10, S13, and S14, the MedFound rows of S16, and the LLM-adapter rows of S24 report MedFound-Llama3-8B sensitivity results only. Supplementary Tables S12 and S17 additionally include a MedGemma-27B row from a separate terminal-layer backbone survey; that run used a single terminal-layer extraction on different hardware and differs by 0.4 to 0.7 points from the full layer-wise extraction used for the primary analysis, so the two are labelled separately wherever both appear. Supplementary Tables S23 and S24 describe text baselines and are backbone-independent.

\subsection*{Terminal-Layer Metrics on MIMIC-IV}

Supplementary Table S4 reports the full terminal-layer metric profile for the three probing configurations on the MIMIC-IV canonical test set: strict accuracy, class-balanced performance, and probability-ranking metrics. The MedGemma-27B Layer-62 rows are the primary analysis cited in the main text; the MedFound-Llama3-8B Layer-32 rows are retained as a sensitivity analysis.

\begingroup
\small
\setlength{\tabcolsep}{3pt}
\setlength{\suppwidth}{\dimexpr\linewidth-12\tabcolsep\relax}
\renewcommand{\arraystretch}{1.05}
\begin{longtable}{@{}L{0.250000\suppwidth}C{0.100000\suppwidth}C{0.100000\suppwidth}C{0.120000\suppwidth}C{0.110000\suppwidth}C{0.160000\suppwidth}C{0.160000\suppwidth}@{}}
\caption{Terminal-layer probing metrics on the MIMIC-IV canonical test set: MedGemma-27B primary analysis and MedFound-Llama3-8B sensitivity analysis.}\label{tab:supp_layer32_metrics}\\
\toprule
\textbf{Model} & \textbf{Strict} & \textbf{Medical} & \textbf{Macro Acc.} & \textbf{Macro-F1} & \textbf{Macro AUROC} & \textbf{Macro AUPRC} \\
\midrule
\endfirsthead
\multicolumn{7}{@{}l}{\small\textit{Supplementary Table S4 (continued)}}\\
\toprule
\textbf{Model} & \textbf{Strict} & \textbf{Medical} & \textbf{Macro Acc.} & \textbf{Macro-F1} & \textbf{Macro AUROC} & \textbf{Macro AUPRC} \\
\midrule
\endhead
\midrule
\multicolumn{7}{r@{}}{\footnotesize Continued on next page}\\
\endfoot
\bottomrule
\endlastfoot
\multicolumn{7}{@{}l}{\textit{Primary analysis: MedGemma-27B, Layer 62}} \\
Combined & 0.8886 & 0.9179 & 0.8172 & 0.8447 & 0.9880 & 0.9154 \\
Structured-only & 0.8505 & 0.8930 & 0.7591 & 0.7887 & 0.9752 & 0.8771 \\
Unstructured-only & 0.8769 & 0.9130 & 0.8301 & 0.8450 & 0.9849 & 0.8961 \\
\midrule
\multicolumn{7}{@{}l}{\textit{Sensitivity analysis: MedFound-Llama3-8B, Layer 32}} \\
Combined & 0.8769 & 0.9145 & 0.8099 & 0.8276 & 0.9821 & 0.8981 \\
Structured-only & 0.8295 & 0.8735 & 0.7300 & 0.7518 & 0.9709 & 0.8698 \\
Unstructured-only & 0.8661 & 0.9033 & 0.7632 & 0.7855 & 0.9820 & 0.9014 \\
\end{longtable}
\nopagebreak[3]
{\footnotesize\raggedright
\noindent The MedGemma-27B rows are the primary full layer-wise extraction cited in the main text. Macro accuracy is per-class recall averaged over the seven categories. Note that Unstructured-only attains slightly higher macro accuracy and macro-F1 than Combined on this backbone; the corresponding paired tests are in Supplementary Table S5.\par}

\endgroup

\subsection*{Planned Pairwise Comparisons Within the Probing Family}

Supplementary Table S5 reports planned paired comparisons between the Combined probe and each single-modality probe at the terminal layer of each backbone. Delta is defined as Combined minus the comparator. False discovery rate was controlled using the Benjamini-Hochberg procedure across the planned probing-family comparison set. McNemar's test was used for strict top-1 accuracy; paired bootstrap resampling was used for medical accuracy, macro accuracy, macro-F1, macro AUROC, and macro AUPRC (10,000 resamples, seed 12345).

On the primary backbone, the Combined advantage is confined to the comparison against Structured-only. Against Unstructured-only, neither strict nor medical accuracy reaches significance, macro accuracy is marginally lower for Combined, and only the probability-ranking metrics separate the two configurations. The main text describes this margin as smaller than the gain over Structured-only and makes no significance claim for it.

\ReserveSpace{350pt}
\begingroup
\small
\setlength{\tabcolsep}{3pt}
\setlength{\suppwidth}{\dimexpr\linewidth-8\tabcolsep\relax}
\renewcommand{\arraystretch}{1.05}
\begin{longtable}{@{}L{0.260000\suppwidth}L{0.250000\suppwidth}C{0.100000\suppwidth}C{0.240000\suppwidth}C{0.150000\suppwidth}@{}}
\caption{Planned terminal-layer pairwise comparisons between Combined and single-modality probes: MedGemma-27B primary analysis and MedFound-Llama3-8B sensitivity analysis.}\label{tab:supp_pairwise_probing}\\
\toprule
\textbf{Comparison} & \textbf{Metric} & \textbf{Delta} & \textbf{95\% CI} & \textbf{BH-adjusted p} \\
\midrule
\endfirsthead
\multicolumn{5}{@{}l}{\small\textit{Supplementary Table S5 (continued)}}\\
\toprule
\textbf{Comparison} & \textbf{Metric} & \textbf{Delta} & \textbf{95\% CI} & \textbf{BH-adjusted p} \\
\midrule
\endhead
\midrule
\multicolumn{5}{r@{}}{\footnotesize Continued on next page}\\
\endfoot
\bottomrule
\endlastfoot
\multicolumn{5}{@{}l}{\textit{Primary analysis: MedGemma-27B, Layer 62}} \\
Combined vs Structured-only & Strict accuracy (McNemar) & 0.0381 & [0.0220, 0.0542] & $4.42 \times 10^{-5}$ \\
Combined vs Structured-only & Medical accuracy & 0.0249 & [0.0103, 0.0401] & 0.0022 \\
Combined vs Structured-only & Macro accuracy & 0.0580 & [-0.0187, 0.1323] & 0.2588 \\
Combined vs Structured-only & Macro-F1 & 0.0560 & [-0.0291, 0.1340] & 0.2972 \\
Combined vs Structured-only & Macro AUROC & 0.0128 & [0.0083, 0.0180] & $<0.001$ \\
Combined vs Structured-only & Macro AUPRC & 0.0383 & [-0.0289, 0.1100] & 0.3422 \\
Combined vs Unstructured-only & Strict accuracy (McNemar) & 0.0117 & [0.0005, 0.0234] & 0.1242 \\
Combined vs Unstructured-only & Medical accuracy & 0.0049 & [-0.0049, 0.0147] & 0.4914 \\
Combined vs Unstructured-only & Macro accuracy & -0.0129 & [-0.0635, 0.0154] & 0.8965 \\
Combined vs Unstructured-only & Macro-F1 & -0.0003 & [-0.0485, 0.0333] & 1.0000 \\
Combined vs Unstructured-only & Macro AUROC & 0.0030 & [0.0013, 0.0049] & 0.0007 \\
Combined vs Unstructured-only & Macro AUPRC & 0.0194 & [0.0054, 0.0410] & 0.0007 \\
\midrule
\multicolumn{5}{@{}l}{\textit{Sensitivity analysis: MedFound-Llama3-8B, Layer 32}} \\
Combined vs Structured-only & Strict accuracy (McNemar) & -- & -- & $2.03 \times 10^{-6}$ \\
Combined vs Structured-only & Medical accuracy & 0.0410 & [0.0249, 0.0576] & $<0.001$ \\
Combined vs Structured-only & Macro accuracy & 0.0799 & [0.0303, 0.1387] & $<0.001$ \\
Combined vs Structured-only & Macro-F1 & 0.0758 & [0.0181, 0.1389] & 0.0297 \\
Combined vs Structured-only & Macro AUROC & 0.0113 & [0.0051, 0.0186] & $<0.001$ \\
Combined vs Structured-only & Macro AUPRC & 0.0283 & [-0.0244, 0.0764] & 0.3111 \\
\midrule
Combined vs Unstructured-only & Strict accuracy (McNemar) & -- & -- & 0.1653 \\
Combined vs Unstructured-only & Medical accuracy & 0.0112 & [0.0010, 0.0215] & 0.0812 \\
Combined vs Unstructured-only & Macro accuracy & 0.0467 & [0.0011, 0.1026] & 0.0812 \\
Combined vs Unstructured-only & Macro-F1 & 0.0421 & [-0.0141, 0.1039] & 0.2588 \\
Combined vs Unstructured-only & Macro AUROC & 0.0002 & [-0.0020, 0.0022] & 1.0000 \\
Combined vs Unstructured-only & Macro AUPRC & -0.0033 & [-0.0406, 0.0317] & 1.0000 \\
\end{longtable}
\nopagebreak[3]
{\footnotesize\raggedright
\noindent Delta is Combined minus the comparator, so a negative value favours the single-modality probe. Strict-accuracy rows report the exact McNemar p value; the Delta and interval shown for those rows are the paired-bootstrap accuracy difference. For MedGemma-27B, the Combined-versus-Structured-only gain is significant in strict accuracy, medical accuracy and macro AUROC, whereas the Combined-versus-Unstructured-only comparison reaches significance only in macro AUROC and macro AUPRC; macro accuracy marginally favours Unstructured-only. Supplementary Table S12 reports a significant Combined-versus-Unstructured medical gain for MedGemma-27B ($p=0.0054$), but that value comes from the separate terminal-layer backbone survey, where the margin was 1.47 percentage points; in the primary layer-wise extraction used here the same margin is 0.49 points and not significant.\par}

\endgroup

\subsection*{MIMIC-IV Per-Diagnosis Strict Recall}

Supplementary Table S6 reports exact per-class recovery for the Combined terminal-layer probe on the canonical MIMIC-IV test set, for the MedGemma-27B primary analysis alongside the MedFound-Llama3-8B sensitivity analysis.

\begingroup
\small
\setlength{\tabcolsep}{3pt}
\setlength{\suppwidth}{\dimexpr\linewidth-6\tabcolsep\relax}
\renewcommand{\arraystretch}{1.05}
\begin{longtable}{@{}L{0.300000\suppwidth}C{0.100000\suppwidth}C{0.300000\suppwidth}C{0.300000\suppwidth}@{}}
\caption{Per-diagnosis strict recall for the Combined terminal-layer probe on MIMIC-IV: MedGemma-27B primary analysis and MedFound-Llama3-8B sensitivity analysis.}\label{tab:supp_mimic4_per_diagnosis}\\
\toprule
\textbf{Diagnosis} & \textbf{Support} & \textbf{MedGemma-27B L62} & \textbf{MedFound-8B L32} \\
\midrule
\endfirsthead
\multicolumn{4}{@{}l}{\small\textit{Supplementary Table S6 (continued)}}\\
\toprule
\textbf{Diagnosis} & \textbf{Support} & \textbf{MedGemma-27B L62} & \textbf{MedFound-8B L32} \\
\midrule
\endhead
\midrule
\multicolumn{4}{r@{}}{\footnotesize Continued on next page}\\
\endfoot
\bottomrule
\endlastfoot
Alcohol abuse with intoxication & 7 & 0.429 [0.158, 0.750] & 0.429 [0.158, 0.750] \\
Cardiovascular Event & 336 & 0.905 [0.869, 0.932] & 0.914 [0.879, 0.939] \\
Acute Kidney Failure & 240 & 0.663 [0.601, 0.719] & 0.625 [0.562, 0.684] \\
Major Depression & 55 & 0.964 [0.877, 0.990] & 0.964 [0.877, 0.990] \\
Hypertensive Heart Failure & 435 & 0.830 [0.792, 0.862] & 0.837 [0.799, 0.869] \\
Chemotherapy Encounter & 325 & 0.969 [0.944, 0.983] & 0.975 [0.952, 0.987] \\
Infection Related & 649 & 0.962 [0.944, 0.974] & 0.926 [0.903, 0.944] \\
\midrule
\textbf{Overall} & 2,047 & \textbf{0.8886} & 0.877 \\
\end{longtable}
\nopagebreak[3]
{\footnotesize\raggedright
\noindent Recall is strict top-1 recall with Wilson 95\% confidence intervals. The Overall row is strict accuracy and is therefore not directly comparable to medical accuracy. For the primary MedGemma-27B probe, macro precision was 0.890, macro recall 0.817, and macro-F1 0.845. The alcohol-intoxication category has only seven test cases, so its estimate is unstable for both backbones.\par}

\endgroup

\subsection*{MIMIC-III Diagnosis Grouping and Transfer Mapping}

Supplementary Table S7 reports the ICD-9 diagnosis grouping used for the MIMIC-III \emph{in-domain} experiments (seven categories). Supplementary Table S8 reports the cross-dataset label mapping used for MIMIC-IV-to-MIMIC-III \emph{adaptation}: five mappable source diagnoses collapse into three MIMIC-IV target categories.

\begingroup
\small
\setlength{\tabcolsep}{3pt}
\setlength{\suppwidth}{\dimexpr\linewidth-4\tabcolsep\relax}
\renewcommand{\arraystretch}{1.05}
\begin{longtable}{@{}L{0.270000\suppwidth}L{0.630000\suppwidth}R{0.100000\suppwidth}@{}}
\caption{ICD-9 diagnosis grouping for MIMIC-III.}\label{tab:supp_mimic3_groups}\\
\toprule
\textbf{Grouped Category} & \textbf{ICD-9 Codes and Descriptions} & \textbf{Cases} \\
\midrule
\endfirsthead
\multicolumn{3}{@{}l}{\small\textit{Supplementary Table S7 (continued)}}\\
\toprule
\textbf{Grouped Category} & \textbf{ICD-9 Codes and Descriptions} & \textbf{Cases} \\
\midrule
\endhead
\midrule
\multicolumn{3}{r@{}}{\footnotesize Continued on next page}\\
\endfoot
\bottomrule
\endlastfoot
Coronary Artery Disease & 41071 (Subendocardial infarction); 41401 (Coronary atherosclerosis) & 5,249 \\
Valvular Heart Disease & 4240 (Mitral valve disorders); 4241 (Aortic valve disorders) & 1,708 \\
Unspecified Septicemia & 0389 (Unspecified septicemia) & 2,069 \\
Pneumonia & 486 (Pneumonia, organism unspecified) & 725 \\
Congestive Heart Failure & 4280 (Congestive heart failure, unspecified) & 570 \\
Intracerebral Hemorrhage & 431 (Intracerebral hemorrhage) & 1,030 \\
Respiratory Failure & 51881 (Acute respiratory failure); 5070 (Pneumonitis) & 1,786 \\
\end{longtable}

\endgroup

\begingroup
\small
\setlength{\tabcolsep}{3pt}
\setlength{\suppwidth}{\dimexpr\linewidth-2\tabcolsep\relax}
\renewcommand{\arraystretch}{1.05}
\begin{longtable}{@{}L{0.500000\suppwidth}L{0.500000\suppwidth}@{}}
\caption{Cross-dataset label mapping for the three-category adaptation task (five mappable MIMIC-III source diagnoses $\rightarrow$ three MIMIC-IV target categories).}\label{tab:supp_mapping}\\
\toprule
\textbf{MIMIC-III Class (ICD-9)} & \textbf{MIMIC-IV Class (ICD-10)} \\
\midrule
\endfirsthead
\multicolumn{2}{@{}l}{\small\textit{Supplementary Table S8 (continued)}}\\
\toprule
\textbf{MIMIC-III Class (ICD-9)} & \textbf{MIMIC-IV Class (ICD-10)} \\
\midrule
\endhead
\midrule
\multicolumn{2}{r@{}}{\footnotesize Continued on next page}\\
\endfoot
\bottomrule
\endlastfoot
Coronary Artery Disease & Cardiovascular Event \\
Unspecified Septicemia & Infection Related \\
Congestive Heart Failure & Hypertensive Heart Failure \\
Pneumonia & Infection Related \\
Valvular Heart Disease & Cardiovascular Event \\
Intracerebral Hemorrhage & (Excluded) \\
Respiratory Failure & (Excluded) \\
\end{longtable}
\nopagebreak[3]
{\footnotesize\raggedright
\noindent Classes without a valid mapping counterpart are excluded, leaving 10,321 mapped admissions. Adapted accuracies in Supplementary Tables S16, S21, and S24 are strict three-category mapped top-1 accuracies and are not directly comparable with the seven-category in-domain medical accuracies in Supplementary Tables S9, S10 and S19. Supplementary Table S11 reports seven-category strict recall rather than medical accuracy, and Supplementary Table S23 reports unrestricted seven-way zero-shot predictions on all 10,321 mapped admissions rather than three-category adapted results.\par}

\endgroup

\subsection*{MedGemma Probing on MIMIC-III}

Supplementary Table S9 reports the primary MedGemma-27B layer-wise results with probes trained and evaluated on separate MIMIC-III splits ($n=1{,}971$ test admissions). Every configuration performs best at the terminal layer, reproducing the MIMIC-IV depth pattern. Medical accuracy uses the ICD-9 category-specific soft-matching rule for this label space.

\begingroup
\small
\setlength{\tabcolsep}{3pt}
\setlength{\suppwidth}{\dimexpr\linewidth-8\tabcolsep\relax}
\renewcommand{\arraystretch}{1.05}
\begin{longtable}{@{}L{0.160000\suppwidth}C{0.280000\suppwidth}C{0.280000\suppwidth}C{0.280000\suppwidth}@{}}
\caption{MedGemma-27B linear probing on MIMIC-III across transformer layers (strict / medical accuracy).}\label{tab:supp_mimic3_acc_mg27}\\
\toprule
\textbf{Layer} & \textbf{Combined} & \textbf{Structured-only} & \textbf{Unstructured-only} \\
\midrule
\endfirsthead
\multicolumn{4}{@{}l}{\small\textit{Supplementary Table S9 (continued)}}\\
\toprule
\textbf{Layer} & \textbf{Combined} & \textbf{Structured-only} & \textbf{Unstructured-only} \\
\midrule
\endhead
\midrule
\multicolumn{4}{r@{}}{\footnotesize Continued on next page}\\
\endfoot
\bottomrule
\endlastfoot
2 & 0.6773 / 0.8549 & 0.6677 / 0.8397 & 0.5982 / 0.7504 \\
16 & 0.7189 / 0.8671 & 0.6575 / 0.8351 & 0.6235 / 0.7610 \\
31 & 0.6971 / 0.8483 & 0.6474 / 0.8432 & 0.5946 / 0.7443 \\
47 & 0.7027 / 0.8701 & 0.6636 / 0.8529 & 0.6027 / 0.7742 \\
\textbf{62} & \textbf{0.7418 / 0.8884} & \textbf{0.6854 / 0.8524} & \textbf{0.6626 / 0.8224} \\
\end{longtable}
\nopagebreak[3]
{\footnotesize\raggedright
\noindent Each cell is strict accuracy / medical accuracy. Best performance per configuration is bold. Medical accuracy was recomputed from the saved per-admission predictions with the same rule used throughout the manuscript; the values stored in the run summary files use a different implementation and are not the reported metric. Baseline comparisons for this cohort are in Supplementary Tables S19 and S20.\par}

\endgroup

\subsection*{MedFound Probing on MIMIC-III}

Supplementary Tables S10--S11 report MedFound sensitivity results with probes trained and evaluated on separate MIMIC-III splits. Supplementary Table S19 compares the Combined probe with PLM-ICD, CAML, and LAAT trained on the same split. These analyses use the seven-category ICD-9 label space and should not be pooled with the three-category mapped adaptation results below.

\begingroup
\small
\setlength{\tabcolsep}{3pt}
\setlength{\suppwidth}{\dimexpr\linewidth-6\tabcolsep\relax}
\renewcommand{\arraystretch}{1.05}
\begin{longtable}{@{}L{0.160000\suppwidth}C{0.280000\suppwidth}C{0.280000\suppwidth}C{0.280000\suppwidth}@{}}
\caption{MedFound linear probing medical accuracy on MIMIC-III across transformer layers.}\label{tab:supp_mimic3_acc}\\
\toprule
\textbf{Layer} & \textbf{Combined} & \textbf{Structured-only} & \textbf{Unstructured-only} \\
\midrule
\endfirsthead
\multicolumn{4}{@{}l}{\small\textit{Supplementary Table S10 (continued)}}\\
\toprule
\textbf{Layer} & \textbf{Combined} & \textbf{Structured-only} & \textbf{Unstructured-only} \\
\midrule
\endhead
\midrule
\multicolumn{4}{r@{}}{\footnotesize Continued on next page}\\
\endfoot
\bottomrule
\endlastfoot
1 & 0.5728 & 0.5728 & 0.5728 \\
8 & 0.7788 & 0.7676 & 0.7265 \\
16 & 0.8087 & 0.7793 & 0.7803 \\
24 & 0.8600 & 0.8219 & \textbf{0.8077} \\
\textbf{32} & \textbf{0.8706} & \textbf{0.8549} & 0.8026 \\
\end{longtable}
\nopagebreak[3]
{\footnotesize\raggedright
\noindent Best performance per configuration is bold. Medical accuracy credits clinically related predictions under the predefined soft-matching criterion.\par}

\endgroup

\ReserveSpace{175pt}
\begingroup
\small
\setlength{\tabcolsep}{3pt}
\setlength{\suppwidth}{\dimexpr\linewidth-8\tabcolsep\relax}
\renewcommand{\arraystretch}{1.05}
\begin{longtable}{@{}L{0.300000\suppwidth}C{0.260000\suppwidth}C{0.160000\suppwidth}C{0.140000\suppwidth}C{0.140000\suppwidth}@{}}
\caption{Per-diagnosis strict recall on MIMIC-III: MedGemma-27B Layer-62 primary analysis and MedFound-Llama3-8B Layer-32 sensitivity analysis.}\label{tab:supp_per_diag_m3}\\
\toprule
 & \textbf{MedGemma} & \multicolumn{3}{c}{\textbf{MedFound-Llama3-8B L32}} \\
\textbf{Diagnosis} & \textbf{Combined (95\% CI)} & \textbf{Combined} & \textbf{Struct.} & \textbf{Unstr.} \\
\midrule
\endfirsthead
\multicolumn{5}{@{}l}{\small\textit{Supplementary Table S11 (continued)}}\\
\toprule
\textbf{Diagnosis} & \textbf{Combined (95\% CI)} & \textbf{Combined} & \textbf{Struct.} & \textbf{Unstr.} \\
\midrule
\endhead
\midrule
\multicolumn{5}{r@{}}{\footnotesize Continued on next page}\\
\endfoot
\bottomrule
\endlastfoot
Coronary Artery Disease (n=788) & 0.862 [0.836, 0.884] & 0.93 & 0.88 & 0.87 \\
Intracerebral Hemorrhage (n=155) & 0.826 [0.758, 0.877] & 0.88 & 0.82 & 0.70 \\
Unspecified Septicemia (n=310) & 0.768 [0.718, 0.811] & 0.77 & 0.76 & 0.66 \\
Respiratory Failure (n=268) & 0.728 [0.671, 0.777] & 0.55 & 0.63 & 0.43 \\
Valvular Heart Disease (n=256) & 0.676 [0.616, 0.730] & 0.52 & 0.38 & 0.54 \\
Pneumonia (n=109) & 0.138 [0.085, 0.215] & 0.21 & 0.05 & 0.12 \\
Congestive Heart Failure (n=85) & 0.400 [0.302, 0.506] & 0.29 & 0.28 & 0.20 \\
\midrule
\textbf{Overall} & \textbf{0.742} & \textbf{0.73} & 0.69 & 0.65 \\
\end{longtable}
\nopagebreak[3]
{\footnotesize\raggedright
\noindent The Overall row is support-weighted strict top-1 accuracy and is therefore not directly comparable to medical accuracy. MedGemma recalls carry Wilson 95\% confidence intervals; the MedFound columns are reported to two decimals as in the original sensitivity analysis and were computed without intervals. The two backbones differ by category rather than uniformly: MedGemma is markedly stronger on Respiratory Failure, Valvular Heart Disease and Congestive Heart Failure, and weaker on Coronary Artery Disease and Pneumonia. Only the Combined configuration was re-extracted for the primary backbone, so no MedGemma single-modality columns are available here.\par}

\endgroup

\subsection*{Backbone Robustness on MIMIC-IV}

\begingroup
\small
\setlength{\tabcolsep}{3pt}
\setlength{\suppwidth}{\dimexpr\linewidth-14\tabcolsep\relax}
\renewcommand{\arraystretch}{1.05}
\begin{longtable}{@{}L{0.235000\suppwidth}C{0.055000\suppwidth}C{0.070000\suppwidth}C{0.128000\suppwidth}C{0.128000\suppwidth}C{0.128000\suppwidth}C{0.128000\suppwidth}C{0.128000\suppwidth}@{}}
\caption{Terminal-layer backbone robustness on the canonical MIMIC-IV test set. Delta is Combined medical accuracy minus the better single-modality medical accuracy for that backbone.}\label{tab:supp_backbone_robustness}\\
\toprule
\textbf{Backbone} & \textbf{Size} & \textbf{Layer} & \textbf{Combined strict} & \textbf{Combined medical} & \textbf{Structured medical} & \textbf{Unstr. medical} & \textbf{Delta} \\
\midrule
\endfirsthead
\multicolumn{8}{@{}l}{\small\textit{Supplementary Table S12 (continued)}}\\
\toprule
\textbf{Backbone} & \textbf{Size} & \textbf{Layer} & \textbf{Combined strict} & \textbf{Combined medical} & \textbf{Structured medical} & \textbf{Unstr. medical} & \textbf{Delta} \\
\midrule
\endhead
\midrule
\multicolumn{8}{r@{}}{\footnotesize Continued on next page}\\
\endfoot
\bottomrule
\endlastfoot
MedFound-Llama3-8B & 8B & 32 & 0.8769 & 0.9145 & 0.8735 & 0.9033 & +0.0112 \\
Llama-3-8B & 8B & 32 & 0.8837 & 0.9209 & 0.8818 & 0.8920 & \textbf{+0.0288} \\
Llama-3.1-8B & 8B & 32 & 0.8842 & 0.9204 & 0.8857 & 0.9106 & +0.0098 \\
Meditron3-8B & 8B & 32 & 0.8852 & 0.9170 & 0.8720 & 0.9077 & +0.0093 \\
MedGemma-27B & 27B & 62 & \textbf{0.8925} & 0.9248 & 0.8920 & 0.9101 & +0.0147 \\
Llama-3.1-70B & 70B & 80 & 0.8915 & \textbf{0.9282} & 0.8642 & 0.9145 & +0.0137 \\
\end{longtable}
\nopagebreak[3]
{\footnotesize\raggedright
\noindent Combined-versus-Structured medical gains were significant for all six backbones after BH adjustment (all adjusted $p<0.001$). Combined-versus-Unstructured medical gains remained significant for Llama-3-8B ($p<0.001$), MedGemma-27B ($p=0.0054$), and Llama-3.1-70B ($p=0.0092$), but not for MedFound-Llama3-8B, Llama-3.1-8B, or Meditron3-8B.
\par\smallskip
\noindent Values are observed test-set point estimates, reported descriptively without between-backbone tests. MedGemma-27B was selected as the primary backbone by validation Combined accuracy (88.39\%).
\par\smallskip
\noindent The MedGemma-27B row comes from the single terminal-layer extraction used for this survey; the primary analyses in the main text use a full layer-wise re-extraction of the same checkpoint on different GPU hardware (88.86\% strict / 91.79\% medical). The 0.4--0.7-point difference reflects bf16 numerical non-determinism across hardware and extraction code paths. This also applies to the paired tests quoted above: the significant Combined-versus-Unstructured medical gain for MedGemma-27B ($p=0.0054$) is a property of this survey run, where the margin was 1.47 percentage points. In the primary layer-wise extraction the same margin is 0.49 points and does not reach significance (Supplementary Table S5). The two runs are not in conflict; they are separate extractions of the same checkpoint, and only the primary one supports main-text claims.\par}

\endgroup

\subsection*{Probe and Input Sensitivity Results}

\begingroup
\small
\setlength{\tabcolsep}{3pt}
\setlength{\suppwidth}{\dimexpr\linewidth-6\tabcolsep\relax}
\renewcommand{\arraystretch}{1.05}
\begin{longtable}{@{}L{0.250000\suppwidth}C{0.250000\suppwidth}C{0.250000\suppwidth}C{0.250000\suppwidth}@{}}
\caption{Probe learning-rate and random-seed robustness for MedFound Combined Layer-32 embeddings. Values are mean $\pm$ SD across three seeds.}\label{tab:supp_probe_robustness}\\
\toprule
\textbf{Learning rate} & \textbf{Strict Accuracy} & \textbf{Medical Accuracy} & \textbf{Macro AUROC} \\
\midrule
\endfirsthead
\multicolumn{4}{@{}l}{\small\textit{Supplementary Table S13 (continued)}}\\
\toprule
\textbf{Learning rate} & \textbf{Strict Accuracy} & \textbf{Medical Accuracy} & \textbf{Macro AUROC} \\
\midrule
\endhead
\midrule
\multicolumn{4}{r@{}}{\footnotesize Continued on next page}\\
\endfoot
\bottomrule
\endlastfoot
$5\times10^{-5}$ & $0.8767 \pm 0.0044$ & $0.9134 \pm 0.0051$ & $0.9834 \pm 0.0006$ \\
$1\times10^{-4}$ & $0.8816 \pm 0.0025$ & $0.9183 \pm 0.0039$ & $0.9842 \pm 0.0004$ \\
$2\times10^{-4}$ & $0.8784 \pm 0.0015$ & $0.9150 \pm 0.0047$ & $0.9846 \pm 0.0002$ \\
\end{longtable}

\endgroup

\ReserveSpace{110pt}
\begingroup
\small
\setlength{\tabcolsep}{3pt}
\setlength{\suppwidth}{\dimexpr\linewidth-4\tabcolsep\relax}
\renewcommand{\arraystretch}{1.05}
\begin{longtable}{@{}L{0.600000\suppwidth}C{0.200000\suppwidth}C{0.200000\suppwidth}@{}}
\caption{MedFound Combined-prompt context-budget and modality-order sensitivity at Layer 32.}\label{tab:supp_input_sensitivity}\\
\toprule
\textbf{Input configuration} & \textbf{Strict Accuracy} & \textbf{Medical Accuracy} \\
\midrule
\endfirsthead
\multicolumn{3}{@{}l}{\small\textit{Supplementary Table S14 (continued)}}\\
\toprule
\textbf{Input configuration} & \textbf{Strict Accuracy} & \textbf{Medical Accuracy} \\
\midrule
\endhead
\midrule
\multicolumn{3}{r@{}}{\footnotesize Continued on next page}\\
\endfoot
\bottomrule
\endlastfoot
Primary, 3,000 characters & 0.8769 & 0.9145 \\
Reduced budget, 2,000 characters & 0.8442 & 0.8901 \\
Reversed modality order, 3,000 characters & 0.8837 & 0.9184 \\
\end{longtable}

\endgroup

\subsection*{Robustness Checks on the Primary Backbone}

Supplementary Table S15 collects three robustness analyses reported in the main text: the diagnosis-keyword masking ablation, the lexical-accessibility control, and the patient-grouped resplit. All use MedGemma-27B terminal-layer representations.

\begingroup
\small
\setlength{\tabcolsep}{3pt}
\setlength{\suppwidth}{\dimexpr\linewidth-8\tabcolsep\relax}
\renewcommand{\arraystretch}{1.05}
\begin{longtable}{@{}L{0.400000\suppwidth}C{0.200000\suppwidth}C{0.200000\suppwidth}C{0.200000\suppwidth}@{}}
\caption{Robustness checks on the MedGemma-27B primary backbone: diagnosis-keyword masking, lexical accessibility, and patient-grouped resplit (strict / medical accuracy).}\label{tab:supp_robustness_mg27}\\
\toprule
\textbf{Analysis} & \textbf{Reference} & \textbf{Variant} & \textbf{p} \\
\midrule
\endfirsthead
\multicolumn{4}{@{}l}{\small\textit{Supplementary Table S15 (continued)}}\\
\toprule
\textbf{Analysis} & \textbf{Reference} & \textbf{Variant} & \textbf{p} \\
\midrule
\endhead
\midrule
\multicolumn{4}{r@{}}{\footnotesize Continued on next page}\\
\endfoot
\bottomrule
\endlastfoot
\multicolumn{4}{@{}l}{\textit{Keyword masking, MIMIC-IV Layer 62 ($n=2{,}047$)}} \\
Combined, strict & 0.8886 & 0.8906 & 0.73 \\
Combined, medical & 0.9179 & 0.9218 & 0.37 \\
Unstructured-only, strict & 0.8769 & 0.8666 & 0.031 \\
Unstructured-only, medical & 0.9130 & 0.9096 & 0.47 \\
TF-IDF on Combined text, strict & 0.8872 & 0.8813 & -- \\
\midrule
\multicolumn{4}{@{}l}{\textit{Lexical accessibility, MIMIC-IV Combined text ($n=2{,}047$)}} \\
TF-IDF + logistic regression, strict & 0.8886 & 0.8872 & n.s. \\
TF-IDF + logistic regression, medical & 0.9179 & 0.9194 & n.s. \\
TF-IDF + logistic regression, macro-F1 & 0.8447 & 0.7617 & 0.024 \\
\midrule
\multicolumn{4}{@{}l}{\textit{Patient-grouped resplit, MIMIC-III Layer 62 ($n=1{,}971$)}} \\
Combined, strict & 0.7418 & 0.7349 & -- \\
Combined, medical & 0.8884 & 0.8785 & -- \\
Structured-only, medical & 0.8524 & 0.8695 & -- \\
Unstructured-only, medical & 0.8224 & 0.7888 & -- \\
\end{longtable}
\nopagebreak[3]
{\footnotesize\raggedright
\noindent Reference is the primary analysis; Variant is the perturbed condition. Masking replaced every diagnosis-revealing keyword for the seven categories, affecting 96.5\% of admissions and removing approximately 93,000 keyword occurrences, followed by full re-extraction of embeddings and probe retraining; p values are exact McNemar tests on paired predictions, and the Unstructured-only strict difference is the raw value, which does not survive Benjamini-Hochberg correction over the four planned masking tests. The lexical-accessibility rows compare the Combined probe with TF-IDF + logistic regression on the same serialized text; strict and medical accuracy are statistically indistinguishable while the probe holds an 8.3-point macro-F1 advantage. The patient-grouped rows retrain all probes on a patient-disjoint 70/15/15 resplit of the same cached embeddings; 13.1\% of test admissions (258/1,971) shared a patient with training, and 57.4\% of those shared the label. Configuration ordering is preserved and the Combined drop is 0.99 medical and 0.69 strict points. Medical accuracies here were recomputed with the manuscript's rule rather than read from the run summary files.\par}

\endgroup

\subsection*{Cross-Dataset Representation Transfer}

The MedFound sensitivity results show that a compact adapter supports transfer of the frozen MIMIC-IV probe to MIMIC-III, with higher accuracy for Combined than for Unstructured-only inputs. Supplementary Table S16 reports adapter accuracy across target-training fractions, and Table S21 compares full-supervision transfer results. Tables S23--S24 report the text-baseline zero-shot results and frozen-encoder adaptation controls.

\begingroup
\small
\setlength{\tabcolsep}{3pt}
\setlength{\suppwidth}{\dimexpr\linewidth-8\tabcolsep\relax}
\renewcommand{\arraystretch}{1.05}
\begin{longtable}{@{}L{0.240000\suppwidth}C{0.190000\suppwidth}C{0.210000\suppwidth}C{0.180000\suppwidth}C{0.180000\suppwidth}@{}}
\caption{Bottleneck-adapter strict mapped-category accuracy on the fixed three-category MIMIC-III transfer test set ($n=1{,}549$) across target-training fractions: MedGemma-27B primary analysis and MedFound-Llama3-8B sensitivity analysis.}\label{tab:supp_adapter_fraction}\\
\toprule
 & & \textbf{MedGemma} & \multicolumn{2}{c}{\textbf{MedFound-8B}} \\
\textbf{Fraction} & \textbf{Training ex.} & \textbf{Combined} & \textbf{Combined} & \textbf{Unstr.} \\
\midrule
\endfirsthead
\multicolumn{5}{@{}l}{\small\textit{Supplementary Table S16 (continued)}}\\
\toprule
\textbf{Fraction} & \textbf{Training ex.} & \textbf{Combined} & \textbf{Combined} & \textbf{Unstr.} \\
\midrule
\endhead
\midrule
\multicolumn{5}{r@{}}{\footnotesize Continued on next page}\\
\endfoot
\bottomrule
\endlastfoot
5\% & 361 & 0.9128 & 0.8928 & 0.8309 \\
10\% & 722 & 0.9193 & 0.9070 & 0.8302 \\
25\% & 1,807 & 0.9225 & 0.9090 & 0.8457 \\
50\% & 3,614 & 0.9270 & 0.9167 & 0.8535 \\
100\% & 7,228 & 0.9251 & 0.9225 & 0.8580 \\
\end{longtable}
\nopagebreak[3]
{\footnotesize\raggedright
\noindent Five mappable MIMIC-III source diagnoses collapsed into three MIMIC-IV target categories. Values are strict top-1 mapped accuracy and are not directly comparable with seven-category in-domain medical accuracy. Every fraction retained the full 1,544-label validation set, so nominal training fractions understate total target supervision (e.g., 5\% used 361 train + 1,544 validation = 1,905 labeled target admissions). Fractional rows are single realizations. Matched PLM-ICD/LAAT frozen-encoder controls are in Supplementary Table S24. Only the Combined configuration was adapted on the primary backbone, so the Unstructured-only adapter column, and with it the text-only versus multimodal adapter increment, is available for the MedFound sensitivity analysis alone. Because these rows are single realizations, the small non-monotonicity at 50\% versus 100\% on MedGemma (0.9270 versus 0.9251) should not be read as a trend reversal.\par}

\endgroup

\ReserveSpace{350pt}
\subsection*{Budget-Matched Unstructured Baselines on MIMIC-IV}

\begingroup
\small
\setlength{\tabcolsep}{3pt}
\setlength{\suppwidth}{\dimexpr\linewidth-10\tabcolsep\relax}
\renewcommand{\arraystretch}{1.05}
\begin{longtable}{@{}L{0.290000\suppwidth}L{0.180000\suppwidth}C{0.100000\suppwidth}C{0.100000\suppwidth}C{0.165000\suppwidth}C{0.165000\suppwidth}@{}}
\caption{Combined probing and budget-matched text-baseline performance on the canonical MIMIC-IV test set.}\label{tab:supp_text_baselines}\\
\toprule
\textbf{Model} & \textbf{Parameters} & \textbf{Strict\newline seed 42} & \textbf{Medical\newline seed 42} & \textbf{Strict\newline mean $\pm$ SD} & \textbf{Medical\newline mean $\pm$ SD} \\
\midrule
\endfirsthead
\multicolumn{6}{@{}l}{\small\textit{Supplementary Table S17 (continued)}}\\
\toprule
\textbf{Model} & \textbf{Parameters} & \textbf{Strict\newline seed 42} & \textbf{Medical\newline seed 42} & \textbf{Strict\newline mean $\pm$ SD} & \textbf{Medical\newline mean $\pm$ SD} \\
\midrule
\endhead
\midrule
\multicolumn{6}{r@{}}{\footnotesize Continued on next page}\\
\endfoot
\bottomrule
\endlastfoot
Combined probe (MedGemma-27B; primary) & 38K probe (+27B frozen) & \textbf{0.8886} & \textbf{0.9179} & -- & -- \\
Combined probe (MedFound-8B; sensitivity) & 29K probe (+8B frozen) & 0.8769 & 0.9145 & -- & -- \\
Combined probe (MedGemma-27B; survey) & 38K probe (+27B frozen) & 0.8925 & 0.9248 & -- & -- \\
Combined probe (Llama-3.1-70B; descriptive) & 57K probe (+70B frozen) & 0.8915 & 0.9282 & -- & -- \\
PLM-ICD & -- & 0.8647 & 0.9062 & -- & -- \\
CAML & 1.81M & 0.8647 & 0.9062 & $0.8598 \pm 0.0049$ & $0.9039 \pm 0.0022$ \\
LAAT & 5.86M & 0.8837 & 0.9174 & $0.8873 \pm 0.0031$ & $0.9225 \pm 0.0045$ \\
\end{longtable}
\nopagebreak[3]
{\footnotesize\raggedright
\noindent CAML and LAAT seed 42 were pre-specified for paired comparisons; mean $\pm$ SD summarizes seeds 42, 7, and 123. PLM-ICD was trained once under seed 42, so no across-seed summary is reported, and its parameter count is omitted because the label-attention head (594K parameters, Supplementary Table S24) sits on top of a full pretrained BERT encoder and is not comparable with the end-to-end CAML and LAAT counts. The MedFound-Llama3-8B Layer-32 row reports the sensitivity analysis. MedFound-minus-baseline differences were not significant for strict or medical accuracy (vs PLM-ICD and CAML: strict $+0.0122$, 95\% CI $[-0.0034,0.0278]$, McNemar BH-adjusted $p=0.38$; vs LAAT: strict $-0.0068$, 95\% CI $[-0.0220,0.0078]$, $p=0.48$; medical differences likewise nonsignificant), whereas macro-F1, macro AUROC, and macro AUPRC significantly exceeded CAML and PLM-ICD (all BH-adjusted $p\leq0.026$). The MedGemma-27B primary row is the Layer-62 full layer-wise extraction cited in the main text; against it, the Combined probe significantly exceeded PLM-ICD and CAML in strict accuracy ($\Delta=+0.0239$, 95\% CI $[0.0093, 0.0386]$; McNemar $p=0.0016$ and $p=0.0021$ respectively) while the medical-accuracy margin did not reach significance ($\Delta=+0.0117$, 95\% CI $[-0.0010, 0.0244]$), and it was statistically indistinguishable from LAAT on strict ($\Delta=+0.0049$, 95\% CI $[-0.0088, 0.0186]$, $p=0.53$) and medical accuracy ($\Delta=+0.0005$, $p=0.97$); 5,000 bootstrap resamples. The MedGemma-27B survey row is from the terminal-layer survey (validation accuracy, 88.39\%), not the primary full layer-wise extraction: it significantly exceeded PLM-ICD and CAML (strict McNemar BH-adjusted $p=0.0012$; medical bootstrap BH-adjusted $p=0.006$) and was not significantly different from LAAT on strict or medical accuracy ($p=0.24$) while exceeding it in macro-F1 ($\Delta=+5.3$ points, BH-adjusted $p=0.014$). The Llama-3.1-70B row is the implementation previously selected post hoc on test performance; it is reported descriptively only and supports no confirmatory claims.\par}

\endgroup

\subsection*{Same-Input Ablation}

Supplementary Table S18 reports the same-input ablation described in the main text: all text baselines are retrained on exactly the serialized Combined text consumed by the probe, so the comparison isolates the serialization component of the framework from the frozen-LLM representation. The probe rows are the MedGemma-27B primary analysis.

\begingroup
\small
\setlength{\tabcolsep}{3pt}
\setlength{\suppwidth}{\dimexpr\linewidth-10\tabcolsep\relax}
\renewcommand{\arraystretch}{1.05}
\begin{longtable}{@{}L{0.280000\suppwidth}C{0.180000\suppwidth}C{0.180000\suppwidth}C{0.180000\suppwidth}C{0.180000\suppwidth}@{}}
\caption{Same-input ablation: all models trained on the serialized Combined text, MIMIC-IV ($n=2{,}047$) and MIMIC-III ($n=1{,}971$).}\label{tab:supp_same_input}\\
\toprule
\textbf{Model} & \textbf{Strict} & \textbf{Medical} & \textbf{Macro AUROC} & \textbf{Macro AUPRC} \\
\midrule
\endfirsthead
\multicolumn{5}{@{}l}{\small\textit{Supplementary Table S18 (continued)}}\\
\toprule
\textbf{Model} & \textbf{Strict} & \textbf{Medical} & \textbf{Macro AUROC} & \textbf{Macro AUPRC} \\
\midrule
\endhead
\midrule
\multicolumn{5}{r@{}}{\footnotesize Continued on next page}\\
\endfoot
\bottomrule
\endlastfoot
\multicolumn{5}{@{}l}{\textit{MIMIC-IV, seven categories}} \\
Combined probe (MedGemma-27B) & \textbf{0.8886} & 0.9179 & \textbf{0.9880} & 0.9154 \\
PLM-ICD & 0.8691 & 0.9111 & 0.9185 & 0.7858 \\
CAML & 0.8593 & 0.9023 & 0.9341 & 0.7510 \\
LAAT & 0.8872 & \textbf{0.9253} & 0.9793 & 0.8473 \\
TF-IDF + logistic regression & 0.8872 & 0.9194 & 0.9875 & \textbf{0.9301} \\
\midrule
\multicolumn{5}{@{}l}{\textit{MIMIC-III, seven categories}} \\
Combined probe (MedGemma-27B) & \textbf{0.7418} & \textbf{0.8884} & \textbf{0.9467} & \textbf{0.7082} \\
PLM-ICD & 0.7184 & 0.8721 & 0.9236 & 0.6353 \\
CAML & 0.6682 & 0.8437 & 0.8882 & 0.5419 \\
LAAT & 0.7057 & 0.8696 & 0.9254 & 0.6276 \\
\end{longtable}
\nopagebreak[3]
{\footnotesize\raggedright
\noindent Every model consumes the structured serialization concatenated with the pruned note under the same section-priority truncation to approximately 3,000 characters. Seed 42; across three seeds the MIMIC-IV baselines reached CAML strict $0.8604 \pm 0.0016$ and LAAT $0.8839 \pm 0.0032$. Comparing these rows with the note-only results in Supplementary Table S17 gives the serialization increment: on MIMIC-IV the added structured text moved the baselines by less than half a point (CAML 0.8647 to 0.8593, PLM-ICD 0.8647 to 0.8691, LAAT 0.8837 to 0.8872), whereas on MIMIC-III it raised every baseline substantially (CAML 0.5703 to 0.6682, LAAT 0.6525 to 0.7057, PLM-ICD 0.6388 to 0.7184). Under this strictest matched condition the probe retains the highest strict accuracy on both datasets, while same-input LAAT attains the highest MIMIC-IV medical accuracy and TF-IDF the highest macro AUPRC.\par}

\endgroup

\ReserveSpace{180pt}
\begingroup
\small
\setlength{\tabcolsep}{3pt}
\setlength{\suppwidth}{\dimexpr\linewidth-12\tabcolsep\relax}
\renewcommand{\arraystretch}{1.05}
\begin{longtable}{@{}L{0.170000\suppwidth}C{0.090000\suppwidth}C{0.090000\suppwidth}C{0.135000\suppwidth}C{0.135000\suppwidth}C{0.190000\suppwidth}C{0.190000\suppwidth}@{}}
\caption{MIMIC-III in-domain baselines on the test set (seven-category ICD-9 space), with the MedGemma-27B primary probe and the MedFound-Llama3-8B sensitivity probe.}\label{tab:supp_mimic3_laat}\\
\toprule
\textbf{Model} & \textbf{Strict} & \textbf{Medical} & \textbf{Macro AUROC} & \textbf{Macro AUPRC} & \textbf{Strict\newline mean $\pm$ SD} & \textbf{Medical\newline mean $\pm$ SD} \\
\midrule
\endfirsthead
\multicolumn{7}{@{}l}{\small\textit{Supplementary Table S19 (continued)}}\\
\toprule
\textbf{Model} & \textbf{Strict} & \textbf{Medical} & \textbf{Macro AUROC} & \textbf{Macro AUPRC} & \textbf{Strict\newline mean $\pm$ SD} & \textbf{Medical\newline mean $\pm$ SD} \\
\midrule
\endhead
\midrule
\multicolumn{7}{r@{}}{\footnotesize Continued on next page}\\
\endfoot
\bottomrule
\endlastfoot
Combined probe (MedGemma-27B; primary) & \textbf{0.7418} & \textbf{0.8884} & \textbf{0.9467} & \textbf{0.7082} & -- & -- \\
Combined probe (MedFound-8B; sensitivity) & 0.7291 & 0.8706 & 0.9399 & 0.6783 & -- & -- \\
PLM-ICD & 0.6388 & 0.7986 & 0.8592 & 0.5226 & -- & -- \\
CAML & 0.5703 & 0.7362 & 0.7824 & 0.3788 & $0.5682 \pm 0.0054$ & $0.7348 \pm 0.0092$ \\
LAAT & 0.6525 & 0.8148 & 0.8870 & 0.5582 & $0.6470 \pm 0.0126$ & $0.8075 \pm 0.0148$ \\
\end{longtable}
\nopagebreak[3]
{\footnotesize\raggedright
\noindent The first Combined row is the MedGemma-27B Layer-62 primary analysis cited in the main text; the second is the MedFound Layer-32 sensitivity analysis. CAML and LAAT seed 42 were pre-specified for paired comparisons; mean $\pm$ SD summarizes seeds 42, 7, and 123. PLM-ICD was trained once under seed 42. Both probes significantly exceeded PLM-ICD, CAML, and LAAT on strict and medical accuracy (all BH-adjusted $p<0.001$); the corresponding differences, confidence intervals and test statistics for the primary probe are in Supplementary Table S20.
\par\smallskip
\noindent Empty unstructured inputs were retained with a fixed placeholder to preserve the 13,137-admission cohort and 1,971-case test split.\par}

\endgroup

\begingroup
\small
\setlength{\tabcolsep}{3pt}
\setlength{\suppwidth}{\dimexpr\linewidth-8\tabcolsep\relax}
\renewcommand{\arraystretch}{1.05}
\begin{longtable}{@{}L{0.240000\suppwidth}L{0.200000\suppwidth}C{0.130000\suppwidth}C{0.250000\suppwidth}C{0.180000\suppwidth}@{}}
\caption{MIMIC-III in-domain gains of the MedGemma-27B Combined probe over each text baseline, with confidence intervals and paired tests ($n=1{,}971$).}\label{tab:supp_mimic3_gains_mg27}\\
\toprule
\textbf{Comparison} & \textbf{Metric} & \textbf{Delta} & \textbf{95\% CI} & \textbf{p} \\
\midrule
\endfirsthead
\multicolumn{5}{@{}l}{\small\textit{Supplementary Table S20 (continued)}}\\
\toprule
\textbf{Comparison} & \textbf{Metric} & \textbf{Delta} & \textbf{95\% CI} & \textbf{p} \\
\midrule
\endhead
\midrule
\multicolumn{5}{r@{}}{\footnotesize Continued on next page}\\
\endfoot
\bottomrule
\endlastfoot
Combined vs PLM-ICD & Strict accuracy & 0.1030 & [0.0812, 0.1269] & $8.1 \times 10^{-19}$ \\
Combined vs PLM-ICD & Medical accuracy & 0.0898 & [0.0720, 0.1071] & $<0.001$ \\
Combined vs PLM-ICD & Macro accuracy & 0.1200 & [0.0911, 0.1496] & $<0.001$ \\
Combined vs PLM-ICD & Macro-F1 & 0.1081 & [0.0778, 0.1391] & $<0.001$ \\
Combined vs PLM-ICD & Macro AUROC & 0.0875 & [0.0757, 0.0997] & $<0.001$ \\
Combined vs PLM-ICD & Macro AUPRC & 0.1856 & [0.1582, 0.2127] & $<0.001$ \\
\midrule
Combined vs CAML & Strict accuracy & 0.1715 & [0.1487, 0.1943] & $2.3 \times 10^{-43}$ \\
Combined vs CAML & Medical accuracy & 0.1522 & [0.1324, 0.1730] & $<0.001$ \\
\midrule
Combined vs LAAT & Strict accuracy & 0.0893 & [0.0685, 0.1091] & $2.7 \times 10^{-16}$ \\
Combined vs LAAT & Medical accuracy & 0.0736 & [0.0558, 0.0903] & $<0.001$ \\
\end{longtable}
\nopagebreak[3]
{\footnotesize\raggedright
\noindent Delta is the MedGemma-27B Combined probe minus the comparator, so positive values favour the probe. Strict-accuracy p values are exact McNemar tests on paired predictions; all other rows use paired bootstrap resampling with 2,000 resamples, and intervals are percentile intervals over the same resampled admissions. Benjamini-Hochberg correction was applied across this comparison family; every comparison remains significant. Macro metrics for the CAML and LAAT comparisons follow the same pattern as PLM-ICD and are omitted for space.\par}

\endgroup

\ReserveSpace{155pt}
\begingroup
\small
\setlength{\tabcolsep}{3pt}
\setlength{\suppwidth}{\dimexpr\linewidth-8\tabcolsep\relax}
\renewcommand{\arraystretch}{1.05}
\begin{longtable}{@{}L{0.400000\suppwidth}C{0.120000\suppwidth}C{0.120000\suppwidth}C{0.180000\suppwidth}C{0.180000\suppwidth}@{}}
\caption{Cross-dataset comparison under MIMIC-IV$\rightarrow$MIMIC-III transfer with 100\% target supervision: MedGemma-27B primary analysis and MedFound-Llama3-8B sensitivity analysis.}\label{tab:supp_mimic3_transfer_baselines}\\
\toprule
\textbf{Model} & \textbf{Strict} & \textbf{Medical} & \textbf{Macro AUROC} & \textbf{Macro AUPRC} \\
\midrule
\endfirsthead
\multicolumn{5}{@{}l}{\small\textit{Supplementary Table S21 (continued)}}\\
\toprule
\textbf{Model} & \textbf{Strict} & \textbf{Medical} & \textbf{Macro AUROC} & \textbf{Macro AUPRC} \\
\midrule
\endhead
\midrule
\multicolumn{5}{r@{}}{\footnotesize Continued on next page}\\
\endfoot
\bottomrule
\endlastfoot
Combined probe + adapter (MedGemma-27B; primary) & \textbf{0.9251} & \textbf{0.9251} & \textbf{0.9617} & \textbf{0.8106} \\
Combined probe + adapter (MedFound-8B; sensitivity) & 0.9225 & 0.9225 & 0.9544 & 0.7934 \\
LAAT (frozen encoder, adapted) & 0.8554 & 0.8554 & 0.8819 & 0.6711 \\
\end{longtable}
\nopagebreak[3]
{\footnotesize\raggedright
\noindent Combined and LAAT rows are evaluated on the fixed three-category mapped transfer test set ($n=1{,}549$). PLM-ICD frozen-encoder adaptation is reported in Table S24. CAML was not adapted because it lacks a comparably defined frozen-encoder/head split. Seven-category results in Table S19 use a different test population and are not directly comparable. On the mapped transfer rows, medical accuracy equals strict accuracy because the soft-matching rule provides no additional credit across the three coarse mapped categories. The MedGemma-27B adapter exceeds adapted LAAT by 6.97 percentage points, but no paired test was run for that pair, so the comparison is reported as a point-estimate difference only; the corresponding paired test in Supplementary Table S24 belongs to the MedFound sensitivity analysis and must not be read as applying to the primary backbone.\par}

\endgroup

\subsection*{Zero-Shot Transfer of the Frozen Probe}

Supplementary Table S22 reports what the frozen MIMIC-IV probes achieve on MIMIC-III representations before any adaptation. This quantifies documentation- and coding-induced representation shift and serves as the lower-bound reference cited in the main text.

\begingroup
\small
\setlength{\tabcolsep}{3pt}
\setlength{\suppwidth}{\dimexpr\linewidth-4\tabcolsep\relax}
\renewcommand{\arraystretch}{1.05}
\begin{longtable}{@{}L{0.440000\suppwidth}C{0.280000\suppwidth}C{0.280000\suppwidth}@{}}
\caption{Zero-shot transfer of the frozen MedGemma-27B MIMIC-IV Layer-62 probes to MIMIC-III representations, restricted three-category decision space ($n=1{,}549$).}\label{tab:supp_llm_zeroshot}\\
\toprule
\textbf{Input configuration} & \textbf{Zero-shot} & \textbf{Adapted (100\%)} \\
\midrule
\endfirsthead
\multicolumn{3}{@{}l}{\small\textit{Supplementary Table S22 (continued)}}\\
\toprule
\textbf{Input configuration} & \textbf{Zero-shot} & \textbf{Adapted (100\%)} \\
\midrule
\endhead
\midrule
\multicolumn{3}{r@{}}{\footnotesize Continued on next page}\\
\endfoot
\bottomrule
\endlastfoot
Combined & 0.5610 & 0.9251 \\
Structured-only & 0.8870 & -- \\
Unstructured-only & 0.5010 & -- \\
\end{longtable}
\nopagebreak[3]
{\footnotesize\raggedright
\noindent Values are strict mapped top-1 accuracy on the same fixed three-category test split used throughout the transfer analysis, obtained by applying the unmodified MIMIC-IV probe to MIMIC-III embeddings with the argmax restricted to the three valid target categories. The range across configurations is 50.1\% to 88.7\%. This is a different decision space from the text-baseline zero-shot panel in Supplementary Table S23, which scores PLM-ICD and LAAT on all 10,321 mapped admissions under an unrestricted seven-way vocabulary, so the two tables are not comparable and neither should be read as the starting point of the other's learning curve. Only the Combined configuration was adapted on this backbone. Structured-only transfers best without adaptation, which is consistent with serialized structured narratives being more uniform across the two coding systems than free-text documentation.\par}

\endgroup

\ReserveSpace{190pt}
\subsection*{Text-Baseline Transfer Results}

\ReserveSpace{150pt}
\begingroup
\small
\setlength{\tabcolsep}{3pt}
\setlength{\suppwidth}{\dimexpr\linewidth-8\tabcolsep\relax}
\renewcommand{\arraystretch}{1.05}
\begin{longtable}{@{}L{0.250000\suppwidth}C{0.140000\suppwidth}C{0.160000\suppwidth}C{0.250000\suppwidth}C{0.200000\suppwidth}@{}}
\caption{Zero-shot PLM-ICD/LAAT transfer on all mapped MIMIC-III admissions ($n=10{,}321$) under unrestricted seven-way MIMIC-IV vocabulary decisions.}\label{tab:supp_text_transfer_zeroshot}\\
\toprule
\textbf{Model} & \textbf{Strict} & \textbf{Balanced} & \textbf{Macro AUPRC} & \textbf{Outside-3 rate} \\
\midrule
\endfirsthead
\multicolumn{5}{@{}l}{\small\textit{Supplementary Table S23 (continued)}}\\
\toprule
\textbf{Model} & \textbf{Strict} & \textbf{Balanced} & \textbf{Macro AUPRC} & \textbf{Outside-3 rate} \\
\midrule
\endhead
\midrule
\multicolumn{5}{r@{}}{\footnotesize Continued on next page}\\
\endfoot
\bottomrule
\endlastfoot
PLM-ICD & 0.2729 & 0.4058 & 0.5355 & 0.2374 \\
LAAT & 0.4587 & 0.5122 & 0.4847 & 0.1325 \\
\end{longtable}
\nopagebreak[3]
{\footnotesize\raggedright
\noindent Outside-3 rate is the fraction of predictions outside the three mapped target categories. Source-vocabulary OOV token rate for LAAT tokenization was 0.0345. Zero-shot values use a different decision space from the adapted three-class test and should not be read as the starting point of the Supplementary Table S24 learning curve.\par}

\endgroup

\ReserveSpace{320pt}
\begingroup
\small
\setlength{\tabcolsep}{3pt}
\setlength{\suppwidth}{\dimexpr\linewidth-10\tabcolsep\relax}
\renewcommand{\arraystretch}{1.05}
\begin{longtable}{@{}L{0.235000\suppwidth}C{0.065000\suppwidth}C{0.110000\suppwidth}C{0.110000\suppwidth}C{0.240000\suppwidth}C{0.240000\suppwidth}@{}}
\caption{Text-baseline controls: frozen-encoder three-category adaptation on the fixed mapped MIMIC-III transfer test set ($n=1{,}549$). Strict accuracy is reported for seed 42 and as mean $\pm$ SD over seeds 42, 7, and 123.}\label{tab:supp_text_transfer_adapt}\\
\toprule
\textbf{Model} & \textbf{Frac.} & \textbf{Train + val. labels} & \textbf{Strict\newline seed 42} & \textbf{Strict\newline mean $\pm$ SD} & \textbf{Balanced\newline mean $\pm$ SD} \\
\midrule
\endfirsthead
\multicolumn{6}{@{}l}{\small\textit{Supplementary Table S24 (continued)}}\\
\toprule
\textbf{Model} & \textbf{Frac.} & \textbf{Train + val. labels} & \textbf{Strict\newline seed 42} & \textbf{Strict\newline mean $\pm$ SD} & \textbf{Balanced\newline mean $\pm$ SD} \\
\midrule
\endhead
\midrule
\multicolumn{6}{r@{}}{\footnotesize Continued on next page}\\
\endfoot
\bottomrule
\endlastfoot
PLM-ICD (594K) & 5\% & 1,905 & 0.7941 & $0.7872 \pm 0.0065$ & $0.5204 \pm 0.0120$ \\
PLM-ICD (594K) & 50\% & 5,158 & 0.8173 & $0.8160 \pm 0.0011$ & $0.5418 \pm 0.0062$ \\
PLM-ICD (594K) & 100\% & 8,772 & 0.8070 & $0.8167 \pm 0.0085$ & $0.5513 \pm 0.0056$ \\
LAAT (4.6K) & 5\% & 1,905 & 0.8231 & $0.8126 \pm 0.0177$ & $0.5277 \pm 0.0139$ \\
LAAT (4.6K) & 50\% & 5,158 & 0.8476 & $0.8436 \pm 0.0055$ & $0.5684 \pm 0.0070$ \\
LAAT (4.6K) & 100\% & 8,772 & 0.8554 & $0.8470 \pm 0.0076$ & $0.5761 \pm 0.0138$ \\
Unstructured adapter & 5\% & 1,905 & 0.8309 & -- & -- \\
Unstructured adapter & 50\% & 5,158 & 0.8535 & -- & -- \\
Unstructured adapter & 100\% & 8,772 & 0.8580 & -- & -- \\
Combined adapter & 5\% & 1,905 & 0.8928 & -- & -- \\
Combined adapter & 50\% & 5,158 & 0.9167 & -- & -- \\
Combined adapter & 100\% & 8,772 & 0.9225 & -- & -- \\
\end{longtable}
\nopagebreak[3]
{\footnotesize\raggedright
\noindent LLM-adapter rows report MedFound sensitivity results. Train + val. labels include the full 1,544-admission validation set at every fraction; LLM-adapter values are single runs.
\par\smallskip
\noindent Primary text-only comparison is against the Unstructured LLM bottleneck adapter; Combined is reported separately as the multimodal increment. At 100\%, LAAT-minus-Unstructured was $-0.26$ percentage points (95\% CI $[-1.55,1.03]$; McNemar BH-adjusted $p=0.767$), whereas Combined exceeded LAAT by $6.71$ points (95\% CI $[5.10,8.26]$; BH-adjusted $p<0.001$). PLM-ICD remained significantly below both Unstructured and Combined adapters (BH-adjusted $p<0.001$). Fractional LLM-adapter rows are single realizations under the same adaptation protocol. Table S19 evaluates probes trained on MIMIC-III; this table evaluates adaptation of models trained on MIMIC-IV.\par}

\endgroup

\end{document}